\documentclass[journal]{IEEEtran}

\usepackage[english]{babel}

\usepackage{ifpdf}

\usepackage{cite} 
\usepackage{url}
\usepackage{hyperref}

\ifCLASSINFOpdf
	\usepackage[pdftex]{graphicx}
	\graphicspath{{./figures/}}
\else
	\usepackage[dvips]{graphicx}
	\graphicspath{./figures/}
\fi
\usepackage{color}
\usepackage{pgf, tikz, pgfplots}
\usetikzlibrary{shapes, arrows, automata, plotmarks}
\usetikzlibrary{calc,hobby,decorations}

\usepackage[safe]{tipa}
\usepackage[cmex10]{amsmath}
\usepackage{amsfonts, amssymb, amsthm}
\usepackage{mathrsfs}
\usepackage{dsfont}

\usepackage{orcidlink}

\usepackage{algorithm,algpseudocode}
	\algnewcommand{\LeftComment}[1]{\Statex \(\triangleright\) #1}

\usepackage{enumerate}
\usepackage{multirow}
\usepackage{rotating}
\usepackage{breqn}
\usepackage{soul}
\usepackage{subcaption}
\usepackage{needspace}

\usepackage{amsmath}

\input{./latex_resources/mySymbol.sty}
\input{./latex_resources/pennColors.sty}

\newcommand{\End}[1]{\text{End}(\mathcal{#1})}

\newtheorem{lemma}{\hspace{0pt}\bf Lemma}

\newtheorem{example}{\hspace{0pt}\bf Example}

\newtheorem{theorem}{\hspace{0pt}\bf Theorem}

\newtheorem{definition}{\hspace{0pt}\bf Definition}

\begin{document}

	
\title{
	Convolutional Learning on Multigraphs
	}


\author{Landon Butler\hspace{0.8mm}\orcidlink{0000-0001-9950-7511}, Alejandro~Parada-Mayorga\hspace{0.8mm}\orcidlink{0000-0003-4425-1816},
        and Alejandro Ribeiro\hspace{0.8mm}\orcidlink{0000-0003-4230-9906}
\thanks{All authors contributed equally. The first author acknowledges support by the National Science Foundation Graduate Research Fellowship under Grant No. DGE-2146752.

Landon Butler is with the Department of Electrical Engineering and Computer Science, University of California, Berkeley, CA 94709, USA (e-mail:
landonb@berkeley.edu).

Alejandro~Parada-Mayorga and Alejandro Ribeiro are with the Department of Electrical and Systems Engineering, University of Pennsylvania, Philadelphia, PA 19104 USA (e-mail:
alejopm@seas.upenn.edu; aribeiro@seas.upenn.edu).}}


\markboth{IEEE Transactions on Signal Processing (SUBMITTED)}%
{Shell \MakeLowercase{\textit{et. al.}}: Bare Demo of IEEEtran.cls for Journals}
\maketitle




\begin{abstract}	

Graph convolutional learning has led to many exciting discoveries in diverse areas. However, in some applications, traditional graphs are insufficient to capture the structure and intricacies of the data. In such scenarios, multigraphs arise naturally as discrete structures in which complex dynamics can be embedded. In this paper, we develop convolutional information processing on multigraphs and introduce convolutional multigraph neural networks (MGNNs). To capture the complex dynamics of information diffusion within and across each of the multigraph's classes of edges, we formalize a convolutional signal processing model, defining the notions of signals, filtering, and frequency representations on multigraphs. Leveraging this model, we develop a multigraph learning architecture, including a generalization of selection sampling to reduce computational complexity. The introduced architecture is applied towards optimal wireless resource allocation and a hate speech localization task, offering improved performance over traditional graph neural networks.

\end{abstract}

\IEEEpeerreviewmaketitle




\section{Introduction}


Graphs have enabled the modeling of relational data in diverse scenarios~\cite{sporns2022graph, dijkstra2019networks, bandyopadhyay2010rewiring}, but are insufficient for representing systems with multiple edge types, such as distinct communication avenues in a social system \cite{meng2016interplay} or different 
frequency bands in a wireless communication system~\cite{9072356}. In these settings, multiple graphs are used with the aim of capturing the complex dynamics occurring within and across each of the edge types \cite{PhysRevLett.110.028701}.

The area of graph signal processing (GSP) has provided the tools to enable learning on graphs through the use of convolutional filtering~\cite{zhang2019graph}. However, the GSP framework fails to capture unique flows of information living on multiple graphs that share the same node set. This is a consequence of the fact that GSP tools can only be applied to each graph in parallel with aggregation across the different edge types occurring after filtering \cite{wang2020abstract,10.1145/3511808.3557572,9564196, ke2021joint, wang2021forecasting}. Although this approach can be effective in scenarios with strong independence between each of the graphs, it fails to capture the rich inter-graph diffusive dynamics contained within the underlying system~\cite{PhysRevLett.110.028701,bertagnolli2021diffusion}. In this paper we derive a framework to represent these dynamics using a convolutional signal model, enabling convolutional learning on \textit{multigraphs}.


A multigraph is defined by a set of nodes interconnected by several types of edges\footnote{In the network science community, these have also been referred to as multiplex, multilayer, multiview, and heterogeneous edge networks.}. With a graph constructed for each edge type, multigraphs can be thought of as the composition of many graphs sharing a common node set. Multigraphs have been used to model and analyze financial systems \cite{molina2015multiplex}, brain networks \cite{lim2019discordant}, and immune response \cite{huang2019principles}. 

Many advances have been made to tackle the challenge of learning on multigraphs \cite{zhang2018scalable,qu2017attention} and heterogeneous networks \cite{fu2020magnn,cen2019representation,yu2022multiplex, 10.1145/3292500.3330961} supporting static information. These methods seek to learn embeddings to be used for tasks such as node classification and link prediction, employing learning to weighted contributions of each of the edge types \cite{qu2017attention,zhang2018scalable} or to identify useful pathways of information flow \cite{cen2019representation,yu2022multiplex, fu2020magnn}. 

We are interested in the much less studied task of learning from signals with multigraph support. For the few prior approaches within this space, convolutional filtering has been applied to each edge class in parallel and aggregated through a learned mixing of the filtered outputs \cite{9564196, ke2021joint, wang2021forecasting}. These approaches fail to capture the rich insights that are captured within the heterogeneous diffusions.

In this paper, we derive a convolutional signal processing framework for multigraphs, which we leverage to define multigraph convolutional neural networks (MGNNs). Our main contributions are the following:
\begin{itemize}
\item[\textbf{(C1)}] We formalize a convolutional multigraph signal processing model rooted in a generic Algebraic Signal Processing (ASP) model, defining the notions of signals, filtering, and frequency representations on multigraphs.
\item[\textbf{(C2)}] We introduce a multigraph convolutional neural network architecture that generalizes and extends the processing of information with multiple graphs that share the same node set.
\item[\textbf{(C3)}] We provide numerical evidence of how it is possible to learn representations on multigraphs. We show that in some scenarios, multigraphs arise naturally and can model both coupled and uncoupled information from multiple graphs.
\end{itemize}

The results presented in this paper have relevant implications on how information is represented and modeled in irregular structures, as the introduced family of architectures is naturally suited to handle heterogeneous relational data. More specifically, we highlight the following as the main implications of this work:

\begin{itemize}
\item[\textbf{(I1)}] Convolutional learning on multigraphs is possible, no matter the properties of the individual graphs that constitute the multigraph.
\item[\textbf{(I2)}] In some scenarios, multigraphs allow the extraction of information and structure in ways that ordinary graphs do not. 
\item[\textbf{(I3)}] Convolutional signal processing on multigraphs naturally captures network dynamics that are not properly described through diffusions on ordinary graphs.
\end{itemize}


This paper is organized as follows. In Section II, we define multigraph signal processing, introducing the notion of signals, filters, and spectral representations. Section III leverages multigraph signal processing to introduce  multigraph convolutional neural networks. This new architecture is applied to two sets of numerical experiments in Section IV, corroborating the results of the paper. Finally, in Section V, a discussion and some conclusions are presented.




\section{Multigraph Signal Processing (MSP)}

In this section, we establish the convolutional signal processing model on multigraphs. Before introducing the concepts of signals and convolutional filters, we formally define multigraphs as follows.


\begin{definition}\label{def_multigraph}
	
	A multigraph $M$ is a pair  $\left( \ccalV,  \{ \ccalE_i \}_{i=1}^{m} \right)$, where $\ccalV$ is a set of nodes and $\{ \ccalE_i \}_{i=1}^{m} $ is a collection of edge sets, such that each  pair $(\ccalV, \ccalE_i)$ constitutes a graph with nodes in $\ccalV$ and edges in $\ccalE_i$.
	
\end{definition}

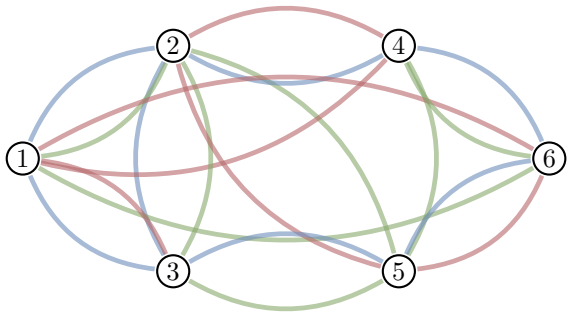
\begin{figure}
		\centering


\definecolor{my_alejocol5}{RGB}{1,31,75}
\definecolor{my_alejocol4}{RGB}{3,57,108}
\definecolor{my_alejocol3}{RGB}{0,91,150}
\definecolor{my_alejocol2}{RGB}{100,151,177}
\definecolor{my_alejocol1}{RGB}{179,205,224}

\colorlet{my_alejocolg1}{black!30}
\colorlet{my_alejocolg2}{black!35}
\colorlet{my_alejocolg3}{black!40}
\colorlet{my_alejocolg4}{black!45}
\colorlet{my_alejocolg5}{black!50}
\colorlet{my_alejocolg6}{black!55}
\colorlet{my_alejocolg7}{black!60}

\definecolor{my_alejocol4s}{RGB}{212, 220, 220}
\definecolor{my_alejocol3s}{RGB}{113, 18, 55}
\definecolor{my_alejocol2s}{RGB}{236, 85, 141}
\definecolor{my_alejocol1s}{RGB}{225, 73, 132}

\definecolor{my_cp4_col1}{RGB}{110,142,187}
\definecolor{my_cp4_col2}{RGB}{182,103,105}
\definecolor{my_cp4_col3}{RGB}{135,169,107}
\definecolor{my_cp4_col4}{RGB}{219,167,108}

\newcommand\DoubleLine[7][4pt]{%
    \path(#2)--(#3)coordinate[at start](h1)coordinate[at end](h2);
    \draw[#4]($(h1)!#1!90:(h2)$)-- node [auto=left] {#5} ($(h2)!#1!-90:(h1)$); 
    \draw[#6]($(h1)!#1!-90:(h2)$)-- node [auto=right] {#7} ($(h2)!#1!90:(h1)$);
    }

  \begin{tikzpicture}[myn/.style={circle,  thick, draw, inner sep=0.05cm, outer sep=1pt, fill=white,opacity=1}]


    \node[myn] (s) at (0,2) {$1$};
    \path (s) + (0,0.8) coordinate (c);

    \node[myn] (a) at (2,3.5) {$2$};
    \path (a) + (0,0.8) coordinate (c);

    \node[myn] (b) at (2,0.5) {$3$};
    \path (b) + (0,-0.8) coordinate (cx);

    \node[myn] (c) at (5,3.5) {$4$};
     \path (c) + (0,0.8) coordinate (cx);

    \node[myn] (d) at (5,0.5) {$5$};
     \path (d) + (0,-0.8) coordinate (cx);

    \node[myn] (t) at (7,2) {$6$};
    \path (t) + (0,0.8) coordinate (cx);

%
%
%


    

    
\path (a) edge  [bend right, line width=1.8, color=my_cp4_col1, opacity=0.6] node {} (b);
\path (a) edge  [bend left, line width=1.8, color=my_cp4_col2, opacity=0.6] node {} (c);
\path (a) edge  [bend left, line width=1.8, color=my_cp4_col3, opacity=0.6] node {} (d);
\path (a) edge  [bend right, line width=1.8, color=my_cp4_col1, opacity=0.6] node {} (s);
\path (t) edge  [bend left, line width=1.8, color=my_cp4_col2, opacity=0.6] node {} (d);
\path (s) edge  [bend right, line width=1.8, color=my_cp4_col3, opacity=0.6] node {} (t);


\path (c) edge  [bend left, line width=1.8, color=my_cp4_col1, opacity=0.6] node {} (t);

\path (c) edge  [bend left, line width=1.8, color=my_cp4_col2, opacity=0.6] node {} (s);

\path (c) edge  [bend left, line width=1.8, color=my_cp4_col3, opacity=0.6] node {} (d);

\path (b) edge  [bend right, line width=1.8, color=my_cp4_col3, opacity=0.6] node {} (d);

\path (b) edge  [bend left, line width=1.8, color=my_cp4_col1, opacity=0.6] node {} (s);


\path (a) edge  [bend left, line width=1.8, color=my_cp4_col3, opacity=0.6] node {} (b);
\path (a) edge  [bend right, line width=1.8, color=my_cp4_col1, opacity=0.6] node {} (c);
\path (a) edge  [bend right, line width=1.8, color=my_cp4_col2, opacity=0.6] node {} (d);
\path (a) edge  [bend left, line width=1.8, color=my_cp4_col3, opacity=0.6] node {} (s);
\path (t) edge  [bend right, line width=1.8, color=my_cp4_col1, opacity=0.6] node {} (d);
\path (s) edge  [bend left, line width=1.8, color=my_cp4_col2, opacity=0.6] node {} (t);

\path (c) edge  [bend right, line width=1.8, color=my_cp4_col3, opacity=0.6] node {} (t);



\path (b) edge  [bend left, line width=1.8, color=my_cp4_col1, opacity=0.6] node {} (d);

\path (b) edge  [bend right, line width=1.8, color=my_cp4_col2, opacity=0.6] node {} (s);

\end{tikzpicture} 
		\caption{Example of a multigraph $M = (\ccalV, \{ \ccalE_1 , \ccalE_2, \ccalE_3 \})$. A set of nodes $\ccalV$ is labeled from 1 to 6. There are three types of edges, $\ccalE_1$, $\ccalE_2$ and $\ccalE_3$, indicated with red, blue and green respectively.}
		\label{fig:basicMultigraph}
\end{figure}

From Definition~\ref{def_multigraph}, we see that a multigraph is an irregular structure, where multiple graph domains are coupled by a set of nodes. Multigraphs allow the modeling of multiple relationships between datapoints -- see Fig.~\ref{fig:basicMultigraph} --
and as we will show later with some examples, in some cases there are non commutative symmetries of data embedded in multigraphs that cannot be captured by a single graph since the traditional signal model for graphs is commutative.


 In order to have matrix representations of a multigraph $M = (\ccalV, \{ \ccalE_i \}_{i=1}^{m})$, we leverage the traditional matrix representations associated to the graphs $\{  (\ccalV, \ccalE_i) \}_{i=1}^{m}$ that comprise $M$. Therefore, if $\bbW_i$ is the adjacency matrix of the graph $(\ccalV, \ccalE_i)$, the multigraph $M$ can be characterized by the collection of adjacency matrices $\{  \bbW_i \}_{i=1}^{m}$. Analogously, we can also use a collection of Laplacian matrices $\{  \bbL_i \}_{i=1}^{m}$ to capture properties of the individual graphs. 
 
  With the basic structure of multigraphs at hand, we are ready to state how information is defined on a multigraph, introducing the notion of signals and filtering in the following subsection.




\begin{figure*}[t]
	\centering
	\begin{subfigure}{.35\textwidth}
		\centering
		\includegraphics[width=1\linewidth]{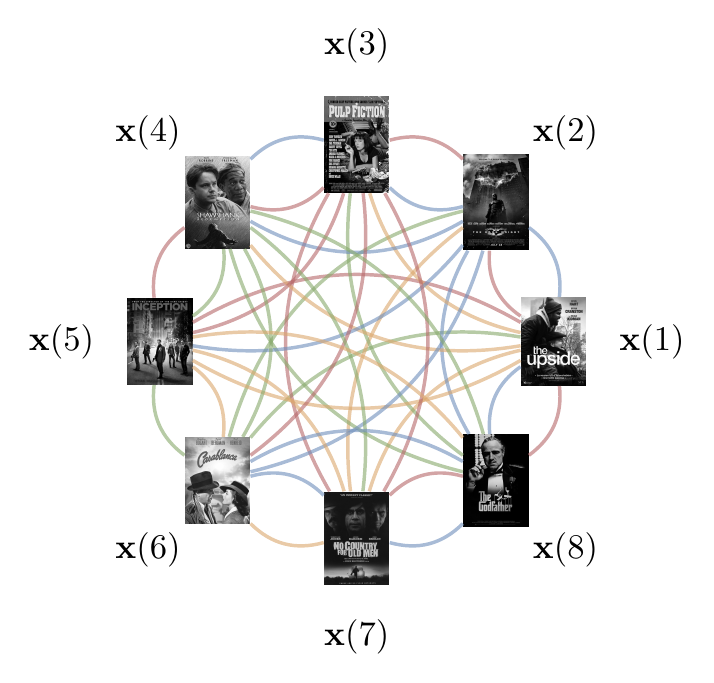} 
		\caption{}
		\label{fig:multiData-one}
	\end{subfigure}
	\begin{subfigure}{.30\textwidth}
		\centering
		\includegraphics[width=1\linewidth]{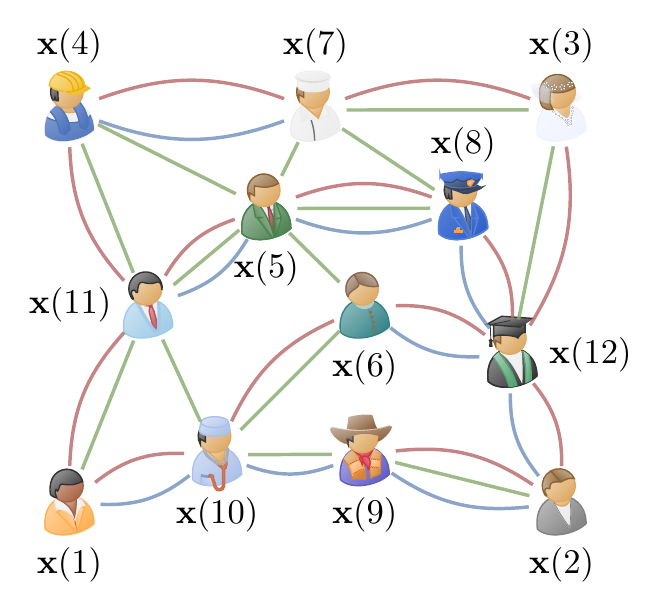} 
		\caption{}
		\label{fig:multiData-two}
	\end{subfigure}
	\begin{subfigure}{.277\textwidth}
		\centering
		\includegraphics[width=1\linewidth]{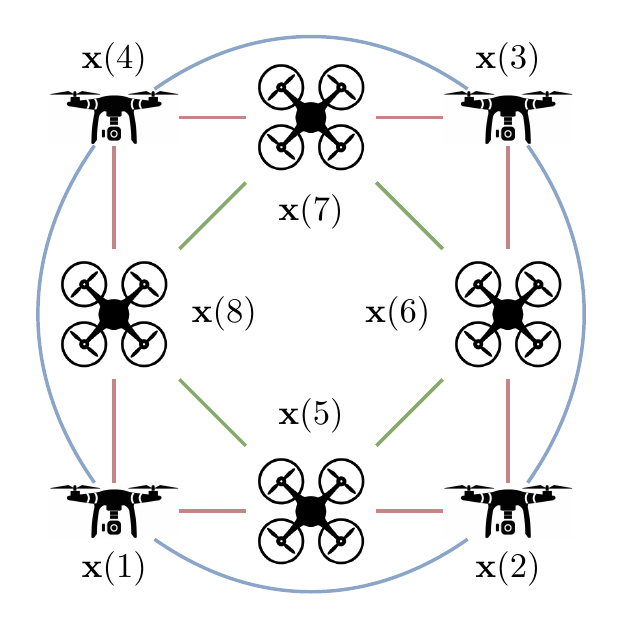} 
		\caption{}
		\label{fig:multiData-two}
	\end{subfigure}
	
	\caption{Examples of signals supported on multigraphs. The signal $\bbx$ has its components defined on the nodes of the multigraph indexed by their numbers. Information is diffused between nodes by means of the different edge types, as indicated by their color. This could represent a recommendation system (a), with the signal storing user ratings and edge types measuring different forms of item similarity. A social network (b) could be modeled, where the signal contains user information and edge types correspond to different relationship types. It could also represent a robotic communication network (c), with the signal storing sensor information and edge types indicating different communication bands.}
	\label{fig:multigraphExamples}
\end{figure*}



\subsection{Signals on Multigraphs and Convolutional Filtering}

In graph signal processing (GSP), signals on graphs are defined as maps from the set of vertices into $\mbR$ and are identified with vectors in $\mbR^{\vert \ccalV \vert}$. Since multigraphs are constituted by graphs sharing a common set of nodes, we introduce the notion of multigraph signals extending the notion of graph signals in the following definition.


\begin{definition}

	Let $M=\left( \ccalV,  \{ \ccalE_i \}_{i=1}^{m} \right)$ be a multigraph. Then, a signal $\bbx$ on $M$
	 is a map $\bbx: \ccalV \rightarrow \mbR$.  
	 
\end{definition}


We identify a signal $\bbx$ on a multigraph $M=\left( \ccalV,  \{ \ccalE_i \}_{i=1}^{m} \right)$ with a vector in $\mbR^{N}$, where $N = \vert \ccalV \vert$. The $i$th component of $\bbx$, given by $\bbx(i)$, is associated to the node in $\ccalV$ indexed by $i$. In Fig. \ref{fig:multigraphExamples}, examples of signals supported on multigraphs are depicted.

Notice that the addition of two signals $\bbx$ and $\bby$ is given by $\bbz = \bbx +\bby$, where $\bbz(i) = \bbx(i) + \bby(i)$, while the multiplication by a scalar $\alpha\in\mbR$ is given by $\alpha\bbx$ with $(\alpha\bbx)(i) = \alpha\bbx(i)$. Then, the signals on a multigraph belong to a vector space that we will denote by $\ccalM$.

With the definition of signals introduced above, we have stated how information can be supported on a multigraph. Now, in order to build a convolutional signal processing model, we need to introduce the notion of convolutional filtering. In GSP, this task is achieved through leveraging the underlying shift operator. We extend this to multigraphs as follows.


\begin{definition}\label{def_shift_operator_multigraph}

	Let $M=\left( \ccalV,  \{ \ccalE_i \}_{i=1}^{m} \right)$ be a multigraph, and let $\bbS_i$ be the shift operator of the graph $ (\ccalV, \ccalE_i) $. Then, each $\bbS_i$ is a shift operator of $M$.
	
\end{definition}


From Definition~\ref{def_shift_operator_multigraph}, we see that the shift operators in the multigraph are inherited from the individual graphs that constitute it. This does not come as a surprise since the shift operator captures properties of the domain that are decoupled into a collection of graph representations. 

Notice that for an individual graph $(\ccalV, \ccalE_i)$, examples of its shift operator include the adjacency matrix $\bbW_i$ and the Laplacian $\bbL_i$. Therefore, a collection of shift operators in the multigraph $M = (\ccalV, \{ \ccalE_i \}_{i=1}^m)$ can be given by $\{ \bbW_i \}_{i=1}^{m}$, $\{ \bbL_i \}_{i=1}^{m}$, or combinations of $\bbW_i$ and $\bbL_i$. It is important to remark that the way each shift operator $\bbS_i$ is selected is independent for each of the graphs. That is, one could choose $\bbS_1 = \bbW_1$ and $\bbS_i =\bbL_i$ for $i=2,\ldots,m$ for instance.

With the notion of shift operators at hand, we are ready to define multigraph filters. We denote the space of multigraph signals defined on the multigraph $M$ by $\ccalM$, and we designate by $\text{End}(\ccalM)$ the set of linear operators acting from $\ccalM$ onto itself.


\begin{definition}

	Let $M= \left( \ccalV, \{ \ccalE_i \}_{i=1}^{m}  \right)$ be a multigraph with shift operators $\{ \bbS_i \}_{i=1}^{m}$. Then, a multigraph filter on $M$ is a polynomial function $p: \text{End}(\ccalM)^m \rightarrow \text{End}(\ccalM)$ whose independent variables are the shift operators $\bbS_i$, and which is denoted by $p(\bbS_1,\ldots,\bbS_m)$. 
	
\end{definition}


The action of $p(\bbS_1,\ldots,\bbS_m)$ on the multigraph signal $\bbx$ is denoted by $p(\bbS_1,\ldots,\bbS_m)\bbx = \bby$. Notice that this notion is consistent since the image of $p(\bbS_1,\ldots,\bbS_m)$ is in $\text{End}(\ccalM)$. We say that $\bby$ is a filtered version of $\bbx$.

We emphasize that no assumption has been made about the commutativity of the shift operators $\{ \bbS_i \}_{i=1}^{m}$. The composition of the action of two filters $p_1 (\bbS_1,\ldots,\bbS_m)$ and $p_2 (\bbS_1, \ldots, \bbS_m)$ is naturally defined by the product of polynomials $(p_1 p_2)(\bbS_1, \ldots, \bbS_m)$, and the operations of addition and multiplication by scalars also follow from the traditional version of these operations with polynomials. For convenience, in what follows, we denote multigraphs filters by $\bbH (\bbS_1, \ldots, \bbS_m)$ which we will denote as $\bbH$ when the shift operators are clear from the context. 

We remark that the filtering operation of signals on multigraphs is a formal convolution. We will show this in Section~\ref{sec_ASP_model} where we derive multigraph signal processing introduced above as a particular instantiation of a generic convolutional algebraic signal model.

\subsection{Diffusions on Multigraphs}

The diffusion of information in any signal model is associated with the action of compositions of shift operators on a given signal. Additionally, such diffusions are captured by the monomial terms under a polynomial representation of the filters. For instance, in GSP, the diffusions are given by the action of $\bbS^k$, where $\bbS$ is the graph shift operator.

\begin{figure}
\centering
\includegraphics[width=\linewidth]{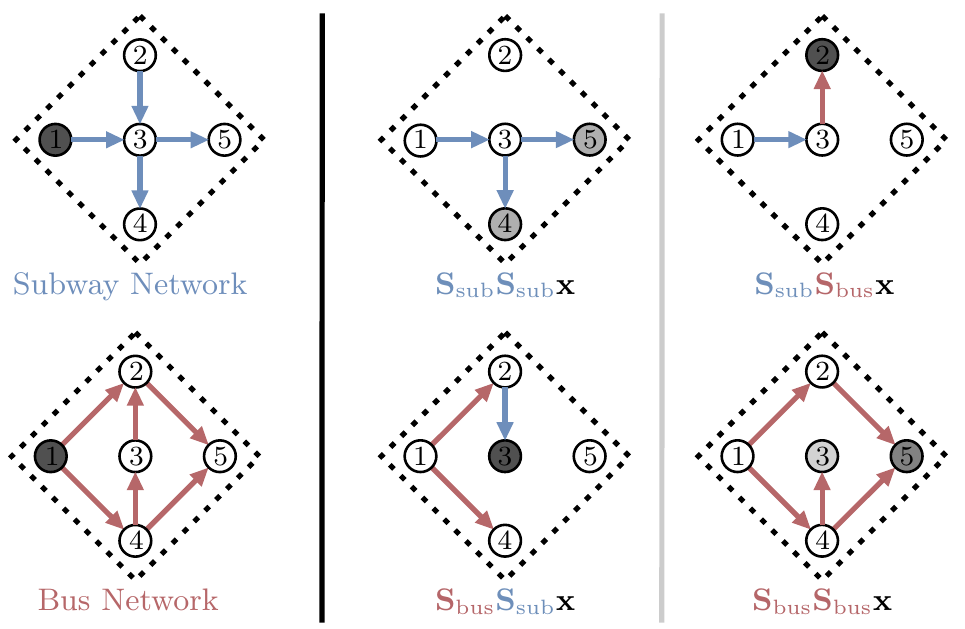}
\caption{Multigraph modeling of a transportation system. Nodes are stations and the edges describe two different methods of transportation between stations. Multigraph signals indicate the amount of passengers in each station. A signal $\bbx$ with $\bbx(i)=0$ for all $i=2,\ldots,5$ describes a scenario where $\bbx(1)$ passengers are at station 1 while all other stations are empty. Left: pictorial description of the individual graphs that constitute the multigraph and the signal $\bbx$. Center and right: all second order diffusions of $\bbx$ are depicted.
%
%
}
\label{fig:urbTrans}
\end{figure}

Multigraph filters are represented by multivariate polynomials, where monomials are given by compositions -- possibly mixed -- of the multiple shift operators. Notice that such compositions are meaningful and interpretable. For instance, consider the multigraph modeling of an urban transportation system, where nodes represent stations, edges denote routes, and edge type corresponds to modality. A passenger may travel using only one modality, but many trips require transferring between buses, subways, or trains. Studying the propagation of passengers across modalities, as shown in Fig. \ref{fig:urbTrans}, allows analysis into the performance of the system, including its resilience  \cite{aleta2017multilayer} and sustainability \cite{natera2020extracting}.

In Fig.~\ref{fig:diffTree}, we depict a tree structure indicating the diffusions associated to multigraph filters with three shift operators $\bbS_i$ for $i=1,2,3$. At the top of the tree, we have the zeroth order diffusion, indicated by the identity operator $\bbI = \bbS_{i}^{0}$, where signals are left identical. In the secondary nodes, we have first order diffusions, where signals are diffused according to $\bbS_i \bbx$ for all $i$, and on the tertiary nodes, we have diffusions given according to $\bbS_i \bbS_j \bbx$ for all $i,j$. The higher the diffusive order, the deeper we traverse down the tree structure and the larger the number of associated monomials. Any multigraph filter can be written as a linear combination of these diffusions. For instance, if we consider only diffusions up to order one, the space of multigraph filters is spanned by $\{\bbI\}\cup\{ \bbS_i \}_{i=1}^{3}$, while if we consider diffusions up to order two, the space of filters is spanned by $\{\bbI\}\cup\{ \bbS_i \}_{i=1}^{3}\cup\{ \bbS_i \bbS_j \}_{i,j=1}^{3}$. 
 
For our analysis in subsequent sections, we characterize multigraph filters $\bbH (\bbS_1,\ldots,\bbS_m)$ in terms of two components. The first component, which we call homogeneous, is associated to univariate polynomials and is given by $\sum_{i=1}^m \bbH_i(\bbS_i)$. The second component, called heterogeneous, relates to the monomials that are multivariate and is represented by $ \bbH_{[1:m]}(\bbS_1,\ldots,\bbS_m) $. Then, we have
\begin{equation}\label{eq_H_decomp_GM}
 \bbH(\bbS_1, \ldots, \bbS_m)
              =
              \sum_{i=1}^{m} \bbH_i (\bbS_i)
              +
              \bbH_{[1:m]}(\bbS_1,\ldots,\bbS_m)
              .
\end{equation}

The representation indicated in~\eqref{eq_H_decomp_GM} is simple, but offers a useful description of how heterogeneous information is naturally embedded and processed in a multigraph. If the size of $\bbH_{[1:m]}(\bbS_1,\ldots,\bbS_m)$ is small -- considering a norm in a space of functions -- we have that information can be filtered independently with filters associated to different interrelation properties of the data. If $\bbH_{[1:m]}(\bbS_1,\ldots,\bbS_m)$ is large, there is a strong, intertwining flow of information between the different layers of the multigraph. We point out that the term $\bbH_{[1:m]}(\bbS_1,\ldots,\bbS_m)$ captures complex dynamics of information that cannot be captured by ordinary graphs.

In what follows, we present an example to provide some intuition about diffusions on multigraphs.


\begin{example}\normalfont

Let us consider Fig.~\ref{fig:urbTrans} where we depict a multigraph $M = (\ccalV, \{ \ccalE_{\text{sub}} , \ccalE_{\text{bus}} \} )$ modeling an urban transportation system. The set of vertices $\ccalV$ models stations and the two sets of edges $\ccalE_{\text{sub}}$ and $\ccalE_{\text{bus}}$ describe paths between stations by subway and bus, respectively. If we denote by $\bbS_{\text{sub}}$ and $\bbS_{\text{bus}}$ the shift operators in the multigraph, convolutional filters and diffusions are defined by compositions of the products, such as $\bbS_{\text{sub}}\bbS_{\text{bus}}$ and $\bbS_{\text{bus}}\bbS_{\text{sub}}$. In Fig.~\ref{fig:urbTrans}, the signal $\bbx$ indicates the number of passengers at each station, with darker nodes containing more passengers. Then, the diffused signals $\bbS_{\text{sub}}\bbS_{\text{bus}}\bbx$ and $\bbS_{\text{bus}}\bbS_{\text{sub}}\bbx$ lead to different destinations within the system. Notice that these particular dynamics cannot be captured alone through compositions of $\bbS_{\text{sub}}^2$ or $\bbS_{\text{bus}}^2$.

\end{example}



\subsection{Operator Pruning}


\begin{figure*}[t]
\centering
\includegraphics[width=\textwidth]{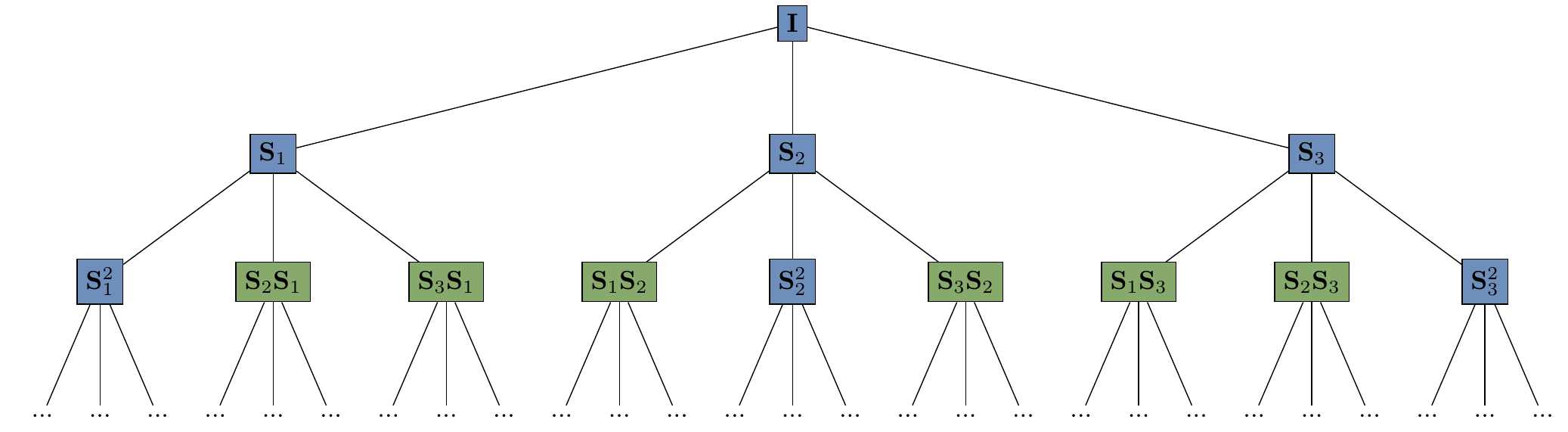}
\caption{Diffusion tree for multigraph filters with three shift operators $\mathbf{S}_1$, $\mathbf{S}_2$, and $\mathbf{S}_3$. Monomials in green are multivariate diffusions, which are not considered by traditional graph neural networks.}
\label{fig:diffTree}
\end{figure*}


The richness of representation provided by multigraph filters comes at a price with respect to the number of monomials associated to each filter. Even with a modest number of shift operators and diffusion order, the number of monomials in the multigraph filter grows dramatically. To see this, let us consider Fig.~\ref{fig:diffTree} where we depict the diffusion structure of multigraph filters with three shift operators. The number of terms associated to diffusions of order one is 3, diffusions of order 2 is 9, and for a diffusion of order $k$ is $3^{k}$. Then, if we have $m$ shift operators, we generate $m^k$ monomial terms for a diffusion of order $k$.

In order to reduce the computational cost, we propose a pruning method that reduces the number of branches considered in the diffusion tree. The proposed method aims to reduce the number of monomials in the filters by grouping those terms that depend on operator compositions that are close to being commutative. This is, if $\Vert \bbS_i \bbS_j - \bbS_j \bbS_i \Vert_2 \leq \epsilon\ll 1$, we build the multigraph filters considering the diffusion tree on those branches that descend from the nodes containing $\bbS_i \bbS_j$ while ignoring those branches descending from nodes containing $\bbS_j \bbS_i$. In Appendix~\ref{app:pruning},  Fig.~\ref{fig:prunedDiffTree}, we depict an example of a pruned tree that follows the approach mentioned above. 

We remark that the proposed approach guarantees a pruning of the branches under the assumption that $\Vert \bbS_i \Vert_2 \leq 1$, which is a common assumption since in most numerical simulations, graph matrices are normalized. This is highlighted in the following lemma.


\begin{lemma}\label{lemma_pruning_bound}

Let $M$ be a multigraph with $m$ shift operators $\{ \bbS_i \}_{i=1}^{m}$, where $\Vert \bbS_i \Vert_2 \leq 1$ for all $i$. Let $[\bbS_i , \bbS_j] = \bbS_i \bbS_j - \bbS_j \bbS_i $ with $\Vert  [\bbS_i , \bbS_j] \Vert_2 \leq \epsilon\ll 1$ for some $i,j$. Then, it follows that

\begin{equation}\label{eq_lemma_pruning_bound_0}
\left\Vert 
      \bbS^{k_1}_{i_1}\cdots\bbS^{k_m}_{i_m}
      [\bbS_i , \bbS_j]
      \bbS_{j_1}^{\ell_1}\cdots\bbS_{j_m}^{\ell_m}
\right\Vert_2
\leq 
\epsilon
\end{equation}
for all $i_r , j_r \in \{ 1,\ldots, m \}$ and $k_r, \ell_r \in\mbN$. 
\end{lemma}

\begin{proof}
The proof follows from the application of the operator norm property and the specific upper bounds for the norms of $\bbS_i$ and $[\bbS_i, \bbS_j]$.
\end{proof}


Lemma~\ref{lemma_pruning_bound} highlights that when two shift operators $\bbS_i$ and $\bbS_j$ are nearly commutative by an error of $\epsilon$, the branches in the diffusion tree associated to terms containing $\bbS_i \bbS_j$ and terms containing $\bbS_j \bbS_i$ differ up to a factor of $\epsilon$. This implies that one can focus on building polynomials considering the branches descending only from terms containing $\bbS_i \bbS_j$. Notice that the branches descending from terms containing  $\bbS_i \bbS_j$ are indeed obtained through taking product between $\bbS_i \bbS_j$ and the terms $\bbS_{i_1}^{k_1}\cdots\bbS_{i_m}^{k_m}$ and $\bbS_{j_1}^{\ell_1}\cdots\bbS_{j_m}^{\ell_m}$ in~\eqref{eq_lemma_pruning_bound_0}. We remark that the proposed approach relies on having a value of $\epsilon\ll 1$, then the pruning method is as accurate as the value of $\epsilon$ allows it. 

We provide the complete procedure for computing the pruned, depth-limited diffusion tree in Alg.~\ref{alg:Pruning} (see Appendix~\ref{app:pruning}).

\subsection{Fourier Analysis on Multigraphs}

In any convolutional signal processing model, the notion of spectral representation or Fourier decomposition is rooted in the notion of irreducibility~\cite{C2,algSP0,algSP1}. While in GSP these concepts translate into spectral decompositions in terms of the eigenvectors of the shift operator, in multigraphs such notion is only valid when all shift operators commute~\cite{C2,algnn_nc_j,algSP0}. As we will show next, Fourier decompositions on multigraphs are expressed in terms of block diagonalizable matrices. This is indeed a consequence of the general fact that irreducible representations of non commutative signal models have a dimension larger than one~\cite{algnn_nc_j}.


\subsubsection{Fourier decomposition of multigraph signals}

Before formally introducing the notion of Fourier decomposition for multigraph signals and multigraph filters, we recall the basics of joint block diagonalization of matrices. 

For a given integer $p\in\mbN$, we define a partition $\tau_p = (p_1, \ldots, p_\ell)$, where $p_j \in \mbN$ and $\sum_{j=1}^{\ell}p_j = p$. Then, given a collection of matrices $\ccalC = \{ \bbC_i \}_{i=1}^{m}$, with $\bbC_i \in\mbR^{N\times N}$, we say that there exists a joint block diagonalization of the elements in $\ccalC$ if there exists a partition $\tau_p$ and a orthonormal matrix $\bbU$ such that
$
\bbC_i 
      =
       \bbU 
           \text{diag}\left( 
                        \boldsymbol{\Sigma}_{1}^{(i)} , \ldots, \boldsymbol{\Sigma}_{\ell}^{(i)}
                       \right) 
       \bbU^{T}
       ,
$       
where $\boldsymbol{\Sigma}_{j}^{(i)}\in\mbR^{p_j \times p_j}$. Notice that each $\boldsymbol{\Sigma}_{j}^{(i)}$ belongs to the space of endomorphisms of a vector space of dimension $p_j$.

With these notions at hand, we formally state the concept of Fourier decomposition of multigraph signals.


\begin{definition}\label{def_Fourier_decomposition_signals}

Let $M$ be a multigraph with shift operators $\{ \bbS_i \}_{i=1}^{m}$, where

\begin{equation}\label{eq_shiftop_blockdiag}
\bbS_i 
      =
       \bbU 
           \textnormal{diag}\left( 
                        \boldsymbol{\Sigma}_{1}^{(i)} , \ldots, \boldsymbol{\Sigma}_{\ell}^{(i)}
                       \right) 
       \bbU^{\intercal}
       ,
\end{equation}
and $\bbU$ is an orthonormal matrix. If $\bbx$ is a signal on $M$, we say that the projection of $\bbx$ on the subspaces associated to $\boldsymbol{\Sigma}_{j}^{(i)}$ is the Fourier decomposition of $\bbx$ and is denoted by $\hat{\bbx}$. In particular, $\hat{\bbx}(j)$ is the Fourier component associated to $\boldsymbol{\Sigma}_{j}^{(i)}$.

\end{definition}


The number $\ell$ of diagonal blocks defines the number of Fourier components of a multigraph signal. We will refer to $\boldsymbol{\Sigma}_{j}^{(i)}$ as the frequencies of the Fourier decomposition. It is worth highlighting that when the shift operators are commutative, we have $\boldsymbol{\Sigma}_{j}^{(i)}\in\mbR^{1\times 1}$, and therefore frequencies are scalars. Then, the Fourier decompositions of the multigraph signals are obtained through projections on 1-dimensional spaces. 
Notice that although the specific choice of the shift operators $\bbS_i$ can be arbitrary between families of edges, some combinations of choices may facilitate or complicate the calculation of the decomposition in~\eqref{eq_shiftop_blockdiag}.

The projection on the space associated to $\Sigma_{j}^{(i)}\in\mbR^{p_j \times p_j}$ is indeed a projection on the related $p_j$ columns of $\bbU$. The expression for $\hat{\bbx}(j)$ is given by $\bbU^{H}_{j}\bbx$,  where $\bbU_j$ is a matrix whose columns are the $p_j$ columns in $\bbU$ and $\hat{\bbx}(j)$ is of dimension $p_j \times 1$. We emphasize that this is a consequence of the non commutativity of the operators and implies that invariances are associated to spaces of dimension larger than one. Then, there are multidimensional subspaces in which information is indistinguishable in terms of invariances. This contrasts with the commutative scenario where invariances are determined by one dimensional spaces.

It is worth pointing out that the spectral decomposition indicated in Definition 5 does not endow the frequencies with an ordering~\cite{jointblock1,jointblock2}. The choice of ordering does not affect the derivation of our results and contributions, but we point out for future research that tools such as total variation measures could be used to define such ordering. This could allow one to have a more physics-oriented interpretation of the frequencies. We recall that total variation measures have been used in GSP to provide an ordering of frequencies defined by complex numbers~\cite{ortega_gsp}.


\subsubsection{Spectral Representation of Multigraph Filters}

Now we turn our attention to the spectral representation of multigraph filters.


\begin{definition}\label{def_filter_spec_rep}
	Let $M$ be a multigraph with shift operators $\{ \bbS_{i}\}_{i=1}^{m}$, where 
$
\bbS_i 
      =
       \bbU 
           \textnormal{diag}\left( 
                        \boldsymbol{\Sigma}_{1}^{(i)} , \ldots, \boldsymbol{\Sigma}_{\ell}^{(i)}
                       \right) 
       \bbU^{T}
       ,
$
	with $\boldsymbol{\Sigma}_{j}^{(i)}\in\mbR^{p_j \times p_j}$, and $\bbU$ orthogonal. If $d = \max_j \{ p_j \}$ and $\boldsymbol{\Lambda}_{i}\in \mbR^{d\times d}$, we say that the polynomial matrix function
	\begin{equation}
	\bbH
	  \left(
	        \boldsymbol{\Lambda}_1, \ldots, \boldsymbol{\Lambda}_m
	  \right)
	  :
	  \left( \mbR^{d\times d} \right)^m \rightarrow \mbR^{d\times d}
	  ,
	\end{equation}
	is the spectral representation of the multigraph filter $\bbH (\bbS_1,\ldots,\bbS_m)$, where $\left( \mbR^{d\times d} \right)^m$ is the $m$-times cartesian product of $\mbR^{d\times d}$.
\end{definition}


 Definition~\ref{def_filter_spec_rep} comes as the natural extension of the spectral representation of filters on ordinary graphs. To see this, we take into account that
\begin{multline}
 \bbH\left( 
       \bbS_1, \ldots, \bbS_m
      \right) 
=
 \bbH\left( 
           \bbU 
               \boldsymbol{\Theta}_1
           \bbU^{\intercal}
           ,
           \bbU 
                \boldsymbol{\Theta}_2
           \bbU^{\intercal}
           ,
           \ldots
           ,
           \bbU
                 \boldsymbol{\Theta}_m
           \bbU^{\intercal}
      \right)
      ,
\end{multline}
where 
$
\boldsymbol{\Theta}_i 
=
\text{diag}\left( 
                  \boldsymbol{\Sigma}_{1}^{(i)} , \ldots, \boldsymbol{\Sigma}_{\ell}^{(i)}
           \right).
$
Since $\bbU$ is orthogonal, it follows that
\begin{multline}
 \bbH\left( 
       \bbS_1, \ldots, \bbS_m
      \right) 
=
 \bbU\text{diag}\left(
     \bbH\left(
             \boldsymbol{\Sigma}_{1}^{(1)} , \ldots, \boldsymbol{\Sigma}_{1}^{(m)}
         \right),
         \ldots
         \right.
         \\
         \left.
         \ldots,
     \bbH\left(
             \boldsymbol{\Sigma}_{\ell}^{(1)} , \ldots, \boldsymbol{\Sigma}_{\ell}^{(m)}
         \right)
         \right)
      \bbU^\intercal
      .
\end{multline}
Then, the term $\bbH(\bbS_1, \ldots, \bbS_m)$ is completely determined by the matrix polynomial functions 
$
 \bbH\left(
             \boldsymbol{\Sigma}_{j}^{(1)} , \ldots, \boldsymbol{\Sigma}_{j}^{(m)}
      \right)
      .
$
Additionally, it is worth pointing out that $\bbH(\bbS_1, \ldots, \bbS_m)\in\mbC^{N\times N}$ with $N = \vert \ccalV \vert$, and 
$
 \bbH\left(
             \boldsymbol{\Sigma}_{j}^{(1)} , \ldots, \boldsymbol{\Sigma}_{j}^{(m)}
      \right)\in\mbC^{p_j \times p_j}
      ,
$
where in general $p_j \ll N$. This is, the functional properties associated to
$
 \bbH\left(
             \boldsymbol{\Sigma}_{j}^{(1)} , \ldots, \boldsymbol{\Sigma}_{j}^{(m)}
      \right)\in\mbC^{p_j \times p_j}
$
are inherited or embedded in 
$
 \bbH\left(
             \bbS_1 , \ldots, \bbS_m
      \right)
      .
$
In~\cite{algnn_nc_j}, it is shown that the properties associated to
$
	\bbH
	  \left(
	        \boldsymbol{\Lambda}_1, \ldots, \boldsymbol{\Lambda}_m
	  \right)
$
with $\boldsymbol{\Lambda}_j \in\mbC^{d\times d}$ determine the properties for any other polynomial $\bbH$ whose independent variables are defined on matrices of dimension lower than $d\times d$. Notice that when the shift operators in the multigraph are commutative, the spectral responses are scalar multivariate functions where each scalar variable is attributed to a classical notion of frequency. 

The relationship between the Fourier decompositions of filters and signals on multigraphs can be summarized as a particular case of the spectral filtering theorem stated in~\cite{algnn_nc_j}, which we adapt here for multigraphs.


\begin{theorem}(Filtering Spectral Theorem~\cite{algnn_nc_j})\label{thm_filtspec}
Let $M$ be a multigraph with shift operators $\{ \bbS_i \}_{i=1}^{m}$. Let
\begin{multline}
 \bbH\left( 
       \bbS_1, \ldots, \bbS_m
      \right) 
=
 \bbU\textnormal{diag}\left(
     \bbH\left(
             \boldsymbol{\Sigma}_{1}^{(1)} , \ldots, \boldsymbol{\Sigma}_{1}^{(m)}
         \right),
         \ldots
         \right.
         \\
         \left.
         \ldots,
     \bbH\left(
             \boldsymbol{\Sigma}_{\ell}^{(1)} , \ldots, \boldsymbol{\Sigma}_{\ell}^{(m)}
         \right)
         \right)
      \bbU^\intercal
\end{multline}
be a multigraph filter where $\bbU$ is orthogonal, and let $\bby = \bbH(\bbS_1, \ldots, \bbS_m)\bbx$. Then, it follows that
\begin{equation}
\hat{\bby}(i)
              =
              \bbH \left( 
                      \boldsymbol{\Sigma}_{i}^{(1)},
                      \ldots,
                      \boldsymbol{\Sigma}_{i}^{(m)}
                 \right)
              \hat{\bbx}(i)  
              ,
\end{equation}
where $\hat{\bby}(i)$ and $\hat{\bbx}(i)$ are the $i$-th components of the Fourier decomposition of $\bby$ and $\bbx$, respectively. 
\end{theorem}


As we mentioned before, the properties of
$
\bbH \left( 
                      \boldsymbol{\Sigma}_{i}^{(1)},
                      \ldots,
                      \boldsymbol{\Sigma}_{i}^{(m)}
                 \right)
$
are inherited from the properties of
     $
     \bbH
	  \left(
	        \boldsymbol{\Lambda}_1, \ldots, \boldsymbol{\Lambda}_m
	  \right)
	  .
     $
Indeed, as indicated in~\cite{algnn_nc_j} when $d>p_j$, one can show that the polynomials
$
\bbH \left( 
           \boldsymbol{\Sigma}_{i}^{(1)},
            \ldots,
            \boldsymbol{\Sigma}_{i}^{(m)}
    \right)
$
satisfy the properties associated to
     $
     \bbH
	  \left(
	        \boldsymbol{\Lambda}_1, \ldots, \boldsymbol{\Lambda}_m
	  \right)
     $
    by completing with zeros. We remark however, that the computation of 
    $
     \bbH\left(
             \boldsymbol{\Sigma}_{\ell}^{(1)} , \ldots, \boldsymbol{\Sigma}_{\ell}^{(m)}
     \right)
    $
    does not require any completion with zeros since the form and coefficients of the polynomial function $\bbH$ are already determined.




\subsection{About the Algebraic Signal Model}\label{sec_ASP_model}

The notions stated above about convolutional information processing on multigraphs are a particular instantiation of convolutional signal models in the light of \textit{algebraic signal processing (ASP)}. Any formal convolutional framework can be obtained as a particular case of a generic algebraic signal model (ASM)~\cite{algSP0, algSP1, algSP2, algSP3}. In ASP, signal models are defined by the triplet 
$
(\ccalA , \ccalM, \rho)
,
$
where $\ccalA$ is an algebra, $\ccalM$ is a vector space on $\mbC$, and $\rho$ is a homomorphism from $\ccalA$ into $\End{M}$ -- the pair $(\ccalM,\rho)$ is a representation of $\ccalA$ in the sense of~\cite{repthybigbook,repthysmbook}. We recall that an algebra $\ccalA$ is a vector space endowed with a product operation that is closed in $\ccalA$, and $\End{M}$ is the space of endomorphisms from $\ccalM$ onto itself. 

For the derivation of the convolutional multigraph signal processing model on the multigraph $M=\left( \ccalV , \{ \ccalE_i \}_{i=1}^{m}\right)$, we choose $\ccalM$ to be the space identified with vectors in $\mbR^{\vert \ccalV \vert}$. If $\bbx\in\ccalM$, the $i$-th component of $\bbx$ given by $ \bbx (i) $ is associated with the $i$-th node in $\ccalV$ -- for a given labeling. The algebra $\ccalA$ is a non commutative polynomial algebra with $m$ generators $\{ t_i \}_{i=1}^{m}$, and $\rho$ is the homomorphism given by $\rho(t_i) = \bbS_i$. 
The multigraph Fourier decomposition is a particular case of the Fourier decomposition for non commutative algebraic models, and each frequency subspace is indeed associated to an irreducible subrepresentation of $(\ccalM,\rho)$ when seen as a representation of $\ccalA$ in the sense specified in~\cite{algnn_nc_j}.




\section{Multigraph Convolutional Neural Networks (MGNNs)}

In this section, we leverage the convolutional signal processing model to present an architecture for convolutional learning on multigraphs. We also discuss dimensionality reduction methods through selection sampling and pooling, and review practical considerations.


\subsection{Multigraph Perceptrons}

We seek to learn a mapping $\mathbf{\Phi}(\cdot)$ between input data $\bbx$ and a target representation $\bby$  parameterized by the underlying structure provided by a multigraph $\mathbf{y} = \mathbf{\Phi}(\bbx;\mathbf{S}_1, \dots, \mathbf{S}_m)$. To achieve this mapping, we can leverage the multigraph convolution $\mathbf{\Phi}(\bbx;\mathbf{S}_1, \dots, \mathbf{S}_m) = \mathbf{H}(\mathbf{S}_1, \dots, \mathbf{S}_m)\bbx $ by learning filter coefficients corresponding to each monomial. To measure the success of $\mathbf{H}$, consider a loss function $J(\cdot)$ and a training set $\mathcal{T}$ consisting of samples $(\bbx, \bby)$. We learn a mapping where $\mathbf{H}^*$ is found through
\begin{equation}
    \label{eq:ERM_MG}
    \mathbf{H}^* = \argmin\limits_{\mathbf{H}}\hspace{0.25em}|\mathcal{T}|^{-1}\sum_{(\bbx,\bby)\in \mathcal{T}} J(\mathbf{y}, \mathbf{H}(\mathbf{S}_1, \dots, \mathbf{S}_m)\bbx).
\end{equation}

Under this formulation, we are learning a linear mapping between  $\mathbf{x}$ and $\mathbf{y}$. For negligible computational increase, we can apply a pointwise nonlinearity $\sigma(\cdot)$ to the output of the multigraph convolution. We refer to this as the \emph{multigraph perceptron}, expressed as $ \sigma(\mathbf{H}(\mathbf{S}_1, \dots, \mathbf{S}_m)\mathbf{x})$. Using the multigraph perceptron, we can parameterize the empirical risk minimization as 
\begin{equation}
    \label{eq:ERM_MG}
    \mathbf{H}^* = \argmin\limits_{\mathbf{H}}\hspace{0.25em}|\mathcal{T}|^{-1}\sum_{(\bbx,\bby)\in \mathcal{T}} J(\mathbf{y}, \sigma(\mathbf{H}(\mathbf{S}_1, \dots, \mathbf{S}_m)\bbx)).
\end{equation}

The multigraph perceptron yields another multigraph signal defined on each node. We recursively build a composition of $L$ multigraph perceptrons, where for each layer $\ell$, we process by
\begin{equation}
    \mathbf{x}_\ell = \sigma(\mathbf{H}_\ell(\mathbf{S}_1, \dots, \mathbf{S}_m)\mathbf{x}_{\ell - 1}).
\end{equation}


\begin{figure}[t]
\centering
\input{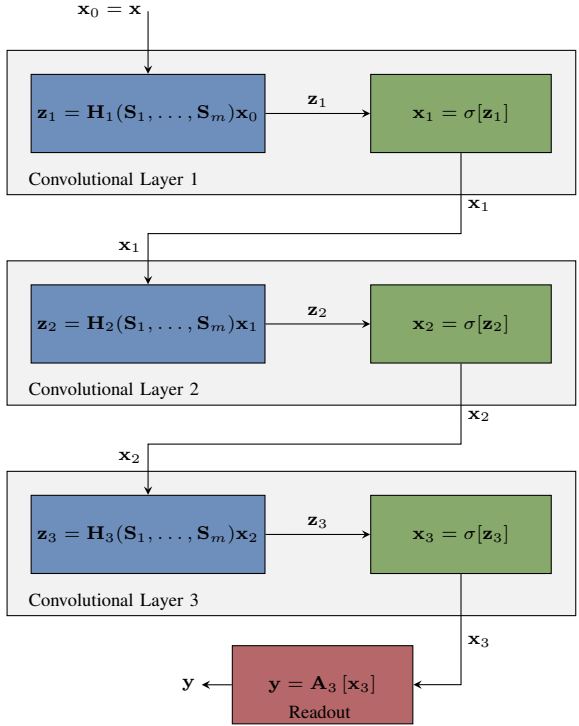}
\caption{Block diagram of a multigraph convolutional neural network with three convolutional layers and a readout.}
\label{fig:blockDiag}
\end{figure}

After the last layer, we can append one or multiple fully connected layers $\mathbf{A}_L$ to map the output of the graph convolutional layers $\mathbf{x}_L$ to the desired shape of $\mathbf{y}$. The concatenation of layers of multigraph perceptrons with layers of readout defines the \emph{multigraph convolutional neural network (MGNN)}. A complete block diagram of this architecture is illustrated in Figure \ref{fig:blockDiag}.

As is the case with graph neural networks (GNNs) \cite{9206091}, multigraph neural networks are equivariant to permutations in node labelings, such that a reordering of the signal components and multigraph shift operators results in the same reordering for the output. Thus, processing by a multigraph
neural network occurs independently of labeling.

Let $\mathbf{P}$ be a permutation matrix of size $N\times N$. When $\mathbf{P}$ is applied to a signal vector $\mathbf{x}_{\ell - 1}$ through $\hat{\mathbf{x}}_{\ell - 1} = \mathbf{P}^\intercal\mathbf{x}_{\ell - 1}$, the resultant signal $\hat{\mathbf{x}}_{\ell - 1}$ is a reordering of the entries of $\mathbf{x}_{\ell - 1}$. Likewise, when $\mathbf{P}$ is applied to each of the shift operators $\{\mathbf{S}_i\}_{i=1}^m$ through $\hat{\mathbf{S}}_i = \mathbf{P}^\intercal\mathbf{S}_i\mathbf{P}$, the rows and columns of each operator are reordered. Together, the permuted shift operators $\{\hat{\mathbf{S}}_i\}_{i=1}^m$ and signals $\hat{\mathbf{x}}_{\ell - 1}$ represent a consistent relabeling of the nodes from the original multigraph. We can now show that MGNNs maintain label permutation equivariance.


\begin{lemma} Given filters $\mathbf{H}_\ell(\mathbf{S}_1, \dots, \mathbf{S}_m)$ and $\mathbf{H}_\ell(\hat{\mathbf{S}}_1, \dots, \hat{\mathbf{S}}_m)$ and corresponding signals $\mathbf{x}_{\ell - 1}$ and $\hat{\mathbf{x}}_{\ell - 1}$, the outputs of each layer $\mathbf{x}_\ell \coloneqq \sigma(\mathbf{H}_\ell(\mathbf{S}_1, \dots, \mathbf{S}_m)\mathbf{x}_{\ell - 1})$ and $\hat{\mathbf{x}}_\ell \coloneqq \sigma(\mathbf{H}_\ell(\hat{\mathbf{S}}_1, \dots, \hat{\mathbf{S}}_m)\hat{\mathbf{x}}_{\ell - 1})$ satisfy
\begin{equation}
\begin{split}
    \hat{\mathbf{x}}_{\ell} :={}&  \sigma(\mathbf{H}_{\ell}(\hat{\mathbf{S}}_1, \dots, \hat{\mathbf{S}}_m)\hat{\mathbf{x}}_{\ell - 1}) \\ :={}& 
    \sigma(\mathbf{H}_{\ell}(\hat{\mathbf{S}}_1, \dots, \hat{\mathbf{S}}_m)(\mathbf{P}^\intercal\mathbf{x}_{\ell - 1})) \\ ={}& \mathbf{P}^\intercal \sigma(\mathbf{H}_{\ell}(\mathbf{S}_1, \dots, \mathbf{S}_m)\mathbf{x}_{\ell - 1}) \\
    :={}& \mathbf{P}^\intercal \mathbf{x}_{\ell}.
    \end{split}
\end{equation}
\end{lemma}

\begin{proof} 
See Appendix \ref{app:permEqui}.
\end{proof}

In convolutional neural networks and graph neural networks, pooling is used to preserve feature information while reducing dimensionality. We generalize pooling to the multigraph setting in the next section.


\subsection{Selection Sampling and Pooling}

For convolutional neural networks, pooling is performed on a regular domain, where homogeneous
geometric information can be used to define a trivial projection onto a lower dimensionality space. This is not the case for graphs nor multigraphs. 

To account for this in the graph setting, coarsening has been used to provide dimensionality reduction \cite{defferrard2016convolutional}. However, it has been shown in \cite{gama2018convolutional} that this process can distort the structure and properties of the original graph. Alternatively, building on the vast GSP literature on sampling by selecting nodes, the authors propose pooling through selection sampling \cite{gama2018convolutional}, in which at each layer, components of the graph signal (or more simply, nodes) are selected that best preserve the graph's original underlying properties and structure. For multigraphs, optimal selection of such nodes remains an open problem \cite{9244194}. For now, we must turn to approximate solutions.

A naive procedure to select nodes would  aggregate across the multigraph (such as by summing the shift operators) and apply a sampling method proposed for graphs \cite{chen2015discrete, anis2016efficient, tsitsvero2016signals, puy2018random, varma2015spectrum}. Under such a method, critical structural properties of the multigraph may be obscured through the aggregation procedure and be unconsidered by the sampling.  To this end, as an approximate solution, \cite{9244194} propose a multigraph generalization of the ``Void and Cluster" (VAC) algorithm intended to capture structural properties within each class of edge, while producing sampling patterns with ideal aggregate properties.

Let $\mathcal{V}_1, \dots, \mathcal{V}_L$ be the set of nodes selected at each layer, with corresponding cardinalities $N_1, \dots, N_L$. These selected nodes maintain the property that if a node is selected at layer $\ell$, it must also be selected for all layers up to $\ell$. When we label nodes such that those which are selected at layer $\ell$ always come before those that are not selected, we can extract the signals of selected nodes of each layer with a $N_{\ell} \times N_{\ell-1}$ wide binary sampling matrix $\mathbf{D}_{\ell}$ containing ones across the main diagonal and zeros elsewhere. We can use this sampling matrix to define new shift operators. Then, the shift operator for the layer $\ell$ and graph $g$ is given by
\begin{equation}
\mathbf{S}_{\ell,g} =\mathbf{D}_{\ell} \mathbf{S}_{\ell-1,g} \mathbf{D}^\intercal_{\ell},
\end{equation}
as well as sampled signals given by
\begin{equation}
\tilde{\mathbf{x}}_{\ell-1} = \mathbf{D}^\intercal_{\ell}\mathbf{x}_{\ell-1},
\end{equation}
where $\mathbf{S}_{0,g} = \mathbf{S}_{g}$ and $\mathbf{x}_{0} = \mathbf{x}$.
With new shift operators and signals on each layer, the output of the multigraph perceptron becomes
\begin{equation}
 \begin{split}
    \mathbf{x}_\ell &= \sigma(\mathbf{H}_\ell(\mathbf{D}_{\ell} \mathbf{S}_{\ell-1,1} \mathbf{D}^\intercal_{\ell}, \dots, \mathbf{D}_{\ell} \mathbf{S}_{\ell-1,m} \mathbf{D}^\intercal_{\ell})\mathbf{D}^\intercal_{\ell}\mathbf{x}_{\ell-1})\\
    &= \sigma(\mathbf{H}_\ell(\mathbf{S}_{\ell,1}, \dots, \mathbf{S}_{\ell,m})\tilde{\mathbf{x}}_{\ell - 1}).\\
 \end{split}
\end{equation}

In order to retain information provided by non-selected nodes, community feature information should be aggregated onto each selected node. This is achieved through pooling.

Let $\alpha_{\ell}$ be a hyperparamter controlling the reach of the pooling operation at each layer. We define $\mathbf{n}_{\ell,i,g}$ to be the $\alpha_{\ell}$-hop neighborhood of node $i$ on graph $g$. At the first layer, this set can be constructed from powers of the shift operators, as this is the set of nodes $j$ with non-zero entries $\left[\mathbf{S}_g^k\right]_{ij} \neq 0$ for all integer powers $k$ up to $\alpha_{\ell}$. For subsequent layers, the neighborhoods of each graph need to be intersected with the set of selected nodes. We can accomplish this through the use of a $N_{\ell} \times N$ wide binary sampling matrix $\mathbf{E}_{\ell}$ also containing ones across the main diagonal and zeros elsewhere
\begin{equation}
    \mathbf{n}_{\ell,i,g} = \left\{j : \left[\mathbf{E}_{\ell}\mathbf{S}_{g}^k\mathbf{E}^\intercal_{\ell - 1}\right]_{ij} \neq 0 \text{ for } k = 0, \dots, \alpha_{\ell} \right\}.
\end{equation}

To define the multigraph neighborhood $ \mathbf{n}_{\ell,i}$ at layer $\ell$, we simply compute the union of each of the graph neighborhoods
\begin{equation}
    \mathbf{n}_{\ell,i} = \bigcup\limits_{g=1}^{m} \mathbf{n}_{\ell,i,g}.
\end{equation}

This set of nodes represents $i$'s locality. When we select nodes at layer $\ell$, if we redefine $i$'s signal as an aggregation over the signals of nodes in $\mathbf{n}_{\ell,i}$, we preserve information of localities. To compute the pooled node signals, we can utilize traditional aggregation operations including the mean, median, or maximum over each feature.

\begin{algorithm}[t]
    \caption{Multigraph (Convolutional) Neural Network (MGNN)}
    \label{alg:selMGNN}
\begin{algorithmic}
\Require  $\mathcal{T}$: training dataset, $\{\hat{\mathbf{x}}\}$: testing dataset, $K$: tree depth,\\ $L$: number of layers, $(\mathbf{S}_1, \dots, \mathbf{S}_m)$: shift operators,\\$\sigma_\ell$: pointwise nonlinearities, $\rho_\ell$: pooling functions, \\$\epsilon$: cutoff parameter, $\texttt{selection}$: selection method 
\Ensure $\{\hat{\mathbf{y}}\}$: predictions of $\{\hat{\mathbf{x}}\}$
\LineComment{\emph{Compute set of operators and instantiate architecture:}}

\If{$\varepsilon == \text{Null}$} 
    \State $\varepsilon \gets \infty$
\EndIf 
\State $\mathcal{Q} = \text{PrunedTree}((\mathbf{S}_1, \dots, \mathbf{S}_m),  \epsilon, K)$

\For{$\ell \gets 1$ to $L$}
\State Compute $\mathbf{D}_{\ell},\mathbf{E}_{\ell}$ from \texttt{selection}

\State Apply pooling $\rho_\ell(\cdot)$ using $\mathbf{D}_{\ell},\mathbf{E}_{\ell}$
\State Sample signal $\bbx_{\ell-1}$ using $\mathbf{D}_{\ell}$ 

\State Create polynomial $\mathbf{H}_\ell(\mathbf{S}_{\ell,1}, \dots, \mathbf{S}_{\ell,m})$ from $\mathcal{Q}$ 

\State Convolve $\mathbf{H}_\ell(\mathbf{S}_{\ell,1}, \dots, \mathbf{S}_{\ell,m})\tilde{\bbx}_{\ell-1}$

\State Apply pointwise nonlinearity $\sigma_\ell(\cdot)$

\EndFor
\State Create fully connected layer $\mathbf{A}_L$
\State
\LineComment{\emph{Train:}}
\State Learn $\{\mathbf{H}_\ell\}_{\ell = 1}^L$ and $\mathbf{A}_L$ from $\mathcal{T}$
\State
\LineComment{\emph{Evaluate:}}
\State Apply MGNN to $\{\hat{\mathbf{x}}\}$ to obtain $\{\hat{\mathbf{y}}\}$
\end{algorithmic}
\end{algorithm}

By leveraging selection and pooling, we retain the structural and signal information necessary to learn informative filters, while reducing the computational expense at each layer. The complete multigraph convolutional neural network (MGNN) algorithm with selection and pooling is provided in Alg. \ref{alg:selMGNN}.


\subsection{Practical Considerations}

\textbf{Multiple-Input-Multiple-Output (MIMO) Multigraph Filters:} The simplest expression of a multigraph filter consists of the application of shift operators weighted by filter coefficients $f$. For instance, an unpruned multigraph filter with three shift operators applied to $\mathbf{x}$ would be expressed as
\begin{equation}
\mathbf{H}(\mathbf{S}_1,\dots,\mathbf{S}_m)\mathbf{x} = f_{\mathbf{I}}\mathbf{I}\mathbf{x} + f_{\mathbf{S}_1}\mathbf{S}_1\mathbf{x} + \dots + f_{\mathbf{S}_1\mathbf{S}_3}\mathbf{S}_1\mathbf{S}_3\mathbf{x} + \dots,
\end{equation}
with learnable coefficients $[f_{\mathbf{I}}, f_{\mathbf{S}_1}, \dots, f_{\mathbf{S}_1\mathbf{S}_3}, \dots]$.

Rather than just considering signals with only one feature, we can accommodate a signal with $F$ features by creating a filter for each feature. Additionally, we can also create a bank of $G$ filters to process each of the $F$ input features. Expressing the multifeature signal as an $N \times F$ matrix $\mathbf{X}$ and filters for each shift operator with an $F \times G$ coefficient matrix $\mathbf{F}$, we can formulate the MIMO multigraph convolution as
\begin{multline}
 \mathbf{H}(\mathbf{S}_1,\dots,\mathbf{S}_m)\mathbf{X} = \\ \mathbf{I}\mathbf{X}\mathbf{F}_{\mathbf{I}} +
\mathbf{S}_1\mathbf{X}\mathbf{F}_{\mathbf{S}_1} + \dots + 
\mathbf{S}_1\mathbf{S}_3\mathbf{X}{\mathbf{F}_{\mathbf{S}_1\mathbf{S}_3}}  + \dots,   
\label{eq:MIMOconv}
\end{multline}
where the output signal is of dimension $N \times G$. Under this formulation, the number of output features at each layer $G_\ell$ is a hyperparameter. The use of multiple features per layer expands the representation power of the MGNN.

\textbf{Scalability:} If no sampling is applied, the cost of the convolution operation at each layer is $\mathcal{O}(N^2m^K F_\ell G_{\ell})$, while the pooled version is $\mathcal{O}(N_{\ell - 1}^2m^K F_\ell G_{\ell})$. When the number of edge classes $m$ is small, this is dominated by the number of nodes at each layer, reinforcing the need for dimensionality reduction through pooling to enable scalability. If $m$ is large, the maximum diffusive order $K$ may need to be reduced, limiting (\ref{eq:MIMOconv}) to low-order polynomials. Though, for graph neural networks, restriction to low-order polynomials is already best practice as a means to prevent over-smoothing \cite{https://doi.org/10.48550/arxiv.2006.13318}.

The number of parameters to be learned at each layer is independent of $N_{\ell - 1}$, as it only depends on the depth of the maximum diffusion order and the number of input and output features: $\mathcal{O}(m^K F_\ell G_{\ell})$.

\textbf{Local Architecture:}
Since the multigraph convolution occurs through local exchanges between a node and its neighbors, the architecture can be implemented in an entirely local manner. This attribute of MGNNs makes this architecture particularly apt for applications in decentralized collaborative systems, in which many multigraphs naturally arise.





%
\section{Applications}
In this section, we compare the proposed MGNN architecture against two other multigraph architectures used in the literature. The first, which we refer to as the parallel graph neural network, processes each edge class by its own graph neural network, and combines the output using a multilayer perceptron \cite{wang2020abstract,10.1145/3511808.3557572}. The second, termed the merged graph neural network, utilizes MIMO convolutions of the form 
\begin{equation}
 \mathbf{H}(\mathbf{S}_1,\dots,\mathbf{S}_m)\mathbf{X} = \mathbf{I}\mathbf{X}\mathbf{F}_{\mathbf{I}} +
 \sum_{i=1}^m \sum_{k = 1}^K \mathbf{S}_i^k\mathbf{X}\mathbf{F}_{\mathbf{S}_i^k}.
\label{eq:mergedConv}
\end{equation}
Note that this differs from \ref{eq:MIMOconv} due to its exclusion of multivariate monomials (heterogeneous diffusions) \cite{9564196, ke2021joint, wang2021forecasting}.

In the first task, we address the problem of optimal resource allocation on a multi-band wireless communication network, where transmitters determine how much power to allot to each frequency band in order to maximize system throughput. For the second task, we address the problem of source localization on a social network, which can be useful for applying interventions to contain the spread of hate speech.


\begin{figure}[t]
\centering
\includegraphics[width=0.7\linewidth]{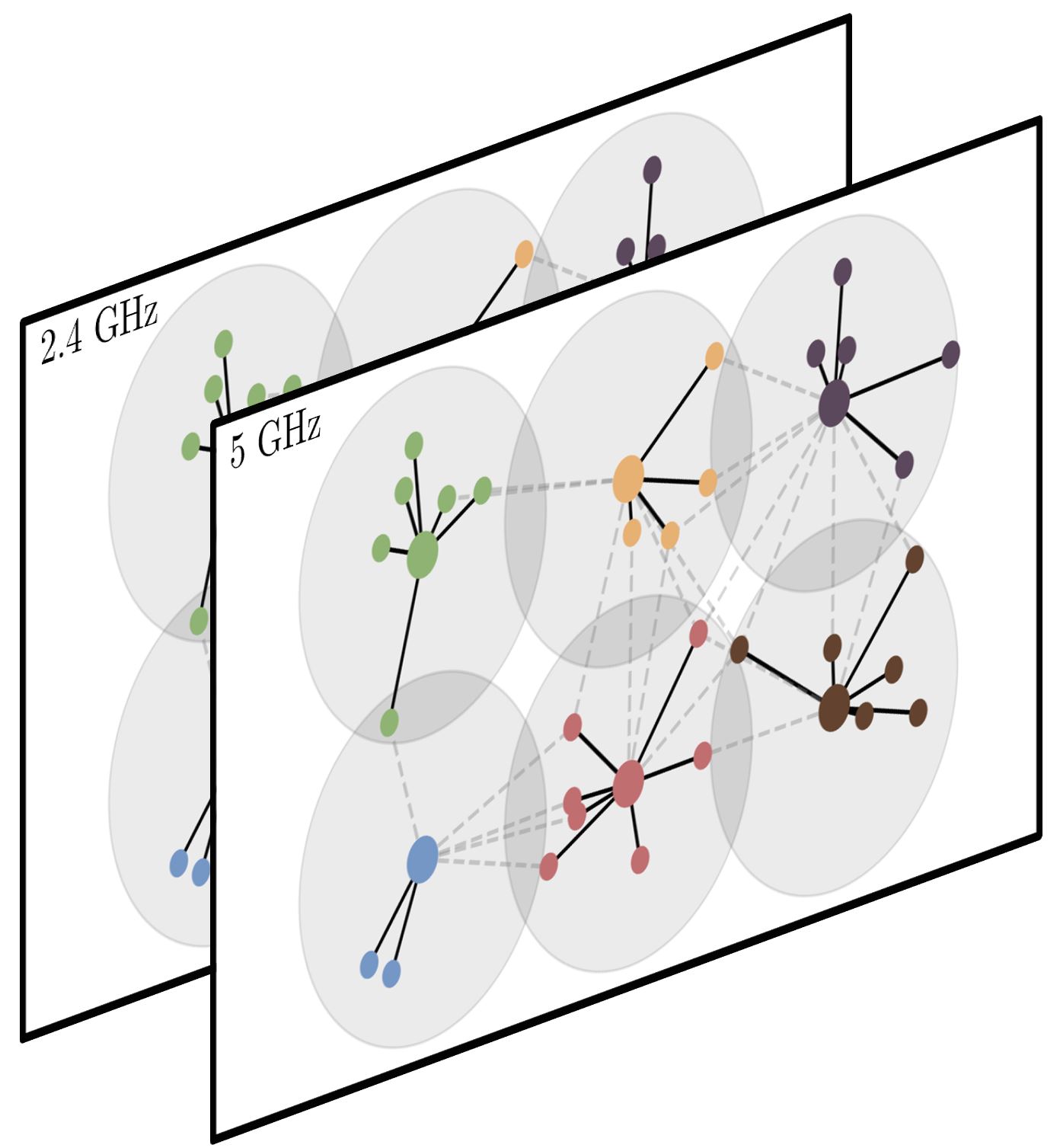}
\caption{For each frequency band, every transmitters competes to communicate with its associated receiver, serving as sources of interference for all other channels.}
\label{fig:wirelessSystem}
\end{figure}



\subsection{Optimal Resource Allocation on Multi-Band Wireless Communication Networks}


\begin{figure*}[t]
	\centering
	\begin{subfigure}{.48\textwidth}
		\centering
		\includegraphics[width=1\linewidth]{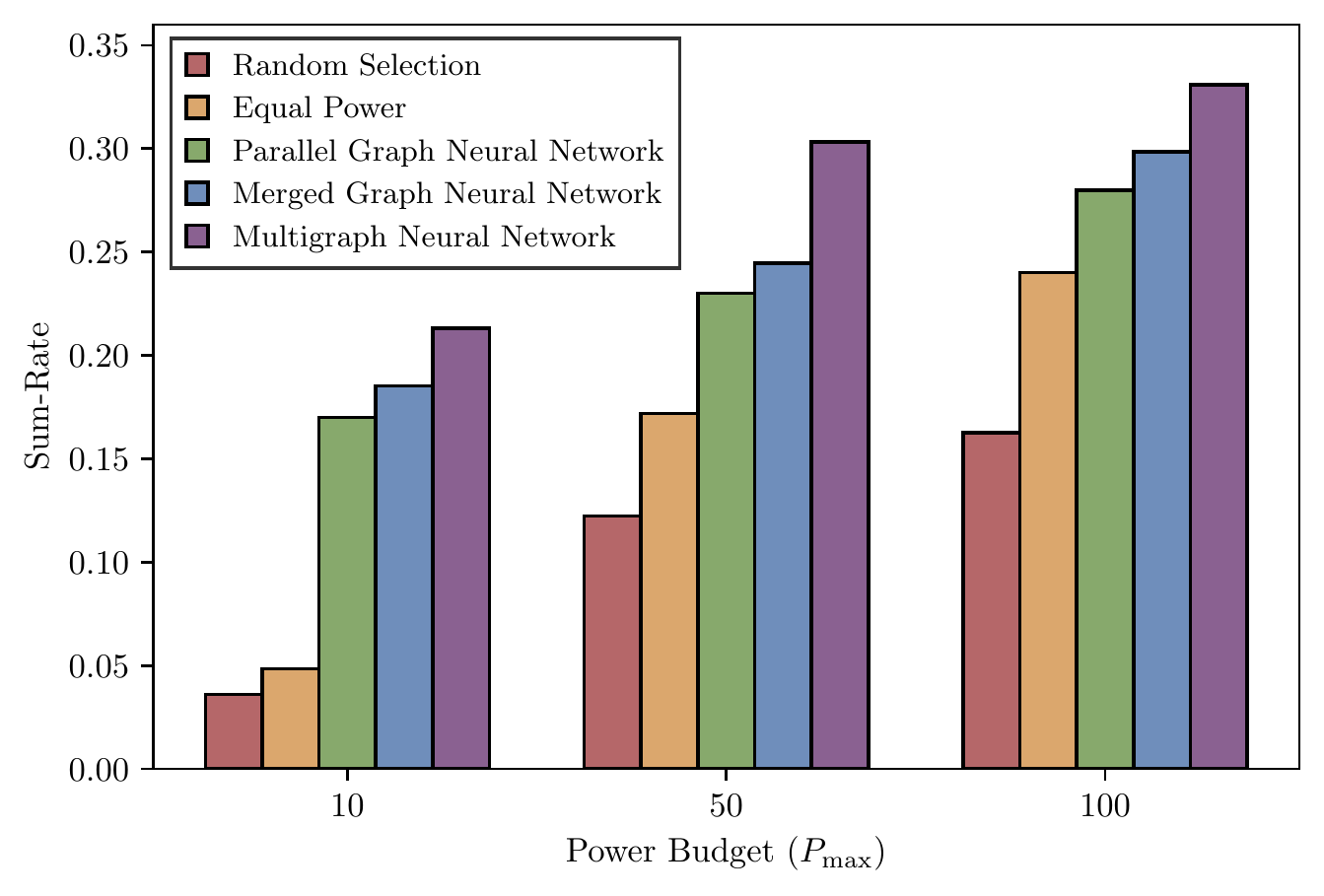} 
		\caption{}
		\label{fig:powerResults}
	\end{subfigure}
	\begin{subfigure}{.48\textwidth}
		\centering
		\includegraphics[width=1\linewidth]{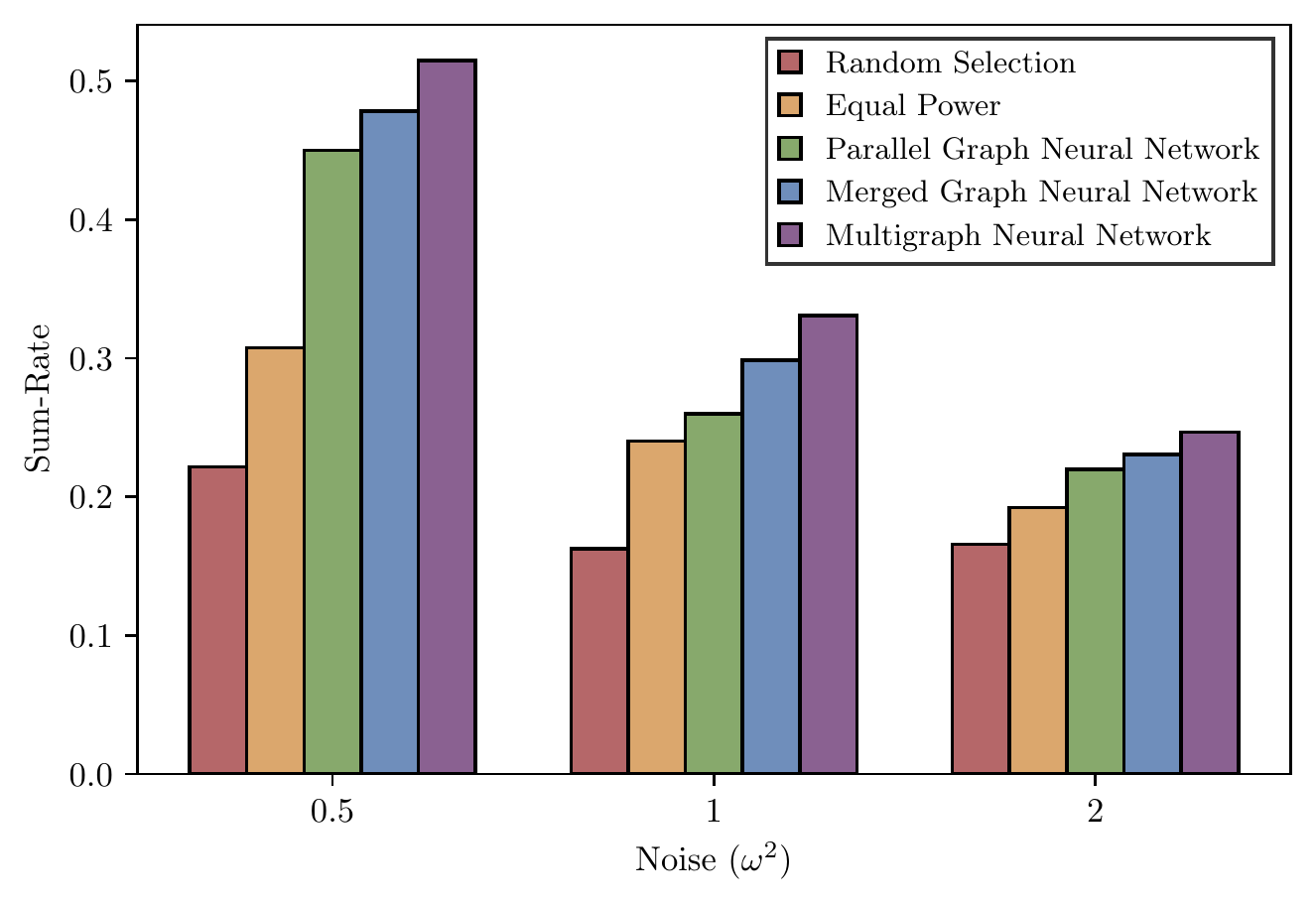} 
		\caption{}
		\label{fig:noiseResults}
	\end{subfigure}

	\caption{Performance comparison of the five architectures.  In (a), the power budget $P_{\text{max}}$ is altered from 10mW to 100mW. For (b), the noise $\omega^2$ is varied, where larger values correspond to lower average SINR. In all regimes, the graph and multigraph neural networks perform much better than the heuristics, with the multigraph neural network offering a boost in performance.}
	\label{fig:wirelessCommPerf}
\end{figure*}


In this task, we seek to maximize the sum-rate of a multi-user interference channel through control of transmit powers. This has been a well studied problem for decades and known to be NP-hard, particularly in the medium to low SIR regime due to its inherent non-convexity. Proposed solutions in the literature approximately solve this problem through exhaustive search \cite{liu2012achieving}, Lyapunov Optimization \cite{8821330}, or iteratively solving approximate subproblems \cite{4275017, chen2011round, shi2011iteratively}. However, when the number of agents in the system is large, these solutions suffer from convergence and scalability issues. 

In response, approximate solutions utilizing machine learning have risen to the forefront in recent years \cite{liang2019towards, 9072356, lima2020resource}, leveraging the scalability of convolutional architectures, making them particularly apt for systems with a large number of agents. First demonstrated in~\cite{9072356}, graph neural networks have been shown to learn effective and scalable policies for optimal wireless resource allocation when all devices share a single frequency band. However, in many settings, devices have the option to transmit within multiple frequency ranges to reduce congestion. Through constructing a communication graph for each frequency band, the entire wireless system can naturally be modeled as a multigraph.

Consider a set of $T$ transmitters and $R$ receivers, where each transmitter is paired with a single receiver. Note that multiple transmitters may be associated with the same receiver, as illustrated in Fig. \ref{fig:wirelessSystem}. Let $b_{i,i}$ represent the channel between a transmitter $i$ and its associated receiver $r(i)$, and let $b_{i,j}$ denote the channel between transmitter $i$ and an unassociated receiver $r(j)$. The channel is composed of $b_{i,j} = b_{i,j}^{pl} b_{i,j}^{f\hspace{-.06cm}f}$, where $b_{i,j}^{pl}$ is a constant path loss gain and $b_{i,j}^{f\hspace{-.06cm}f}$ is the time varying fast fading. We compute the free space path loss $\psi_{i,j}$ through
\begin{equation}\label{eqn:fspl}
\psi_{i,j} = 20\log_{10}(d_{i,j})+20\log_{10} (\nu) + 32.45,
\end{equation}
where $d_{i,j}$ is the distance (in meters) between transmitter $i$ and receiver $r(j)$, and $\nu$ is the frequency (in GHz) \cite{goldsmith2005wireless}.
From path loss, we can define the channel gain as
\begin{equation}\label{eqn:channel-gain}
b_{i,j}^{pl} = 10^{-\frac{\psi_{i,j}}{10}}.
\end{equation}

The fast fading coefficients $b_{i,j}^{f\hspace{-.06cm}f}$ are sampled from a standard Rayleigh distribution. All channels can be coalesced in matrix $\mathbf{B} \in \mbR ^ {T\times T}$ with entries $[\mathbf{B}]_{i,j} = b_{i,j}$. Diagonal elements $b_{i,i}$ represent the channel between each transmitter and its associated receivers. Off-diagonal elements $b_{i,j}$ represent sources of interference for the channel between transmitter $i$ and receiver $r(i)$ caused by communications between transmitter $j$ and receiver $r(j)$. Note that this includes the case when $r(i) = r(j)$, as this is also interference for transmitter $i$'s channel. We sparsify $\mathbf{B}$ by keeping the 20 largest entries for each row, and zeroing out the rest.

For our experiment, we permit transmitters to communicate over the 2.4GHz and 5GHz frequency bands, such that $\nu \in \{2.4, 5\}$. Since each band induces different channel strength due to free space path loss, we denote the channel matrix by $\mathbf{B}^{(\nu)}$ containing elements $b_{i,j}^{(\nu)}$. Each transmitter selects how much power they wish to allocate to  each band through $q_i^{(\nu)}$, which can be grouped into $\mathbf{q}^{(\nu)} \in \mbR_{\geq 0}^T$. 

To measure system performance, we compute the capacity $c_i^{(\nu)}$ experienced by each channel between transmitter $i$ and its receiver $r(i)$ for frequency $\nu$ under additive white Gaussian noise $\omega^2$ and multi-user interference through

\begin{equation}
c_i^{(\nu)}(\mathbf{q}^{(\nu)},\mathbf{B}^{(\nu)}) = \mbE \left[\log \left(1 + \frac{b_{i,i}^{(\nu)}q_i^{(\nu)}}{\omega^2 + \sum_{j\neq i}b_{j,i}^{(\nu)}q_j^{(\nu)}}\right)\right]
,
\end{equation}
where $\mbE$ is the expected value operator.

All transmitters are subject to a shared maximum average power budget $P_{\text{max}}$. With this constraint, we seek to maximize the expected sum-rate (i.e. the expected sum of all capacities over both frequency bands). The resource allocation problem can be formulated as

\begin{equation}
\begin{aligned}
\max_{\mathbf{q}} \quad & \sum_{\nu \in \{2.4,5\}}\sum_{i=1}^T c_i^{(\nu)}(\mathbf{q}^{(\nu)},\mathbf{B}^{(\nu)})\\\
\textrm{s.t.} \quad & \sum_{\nu \in \{2.4,5\}}\mbE[\mathbf{1}^\intercal \mathbf{q}^{(\nu)}] \leq P_{\text{max}}\\
  & \mathbf{q}^{(\nu)} \in  \mbR_{\geq 0}^T \quad \nu \in \{2.4, 5\} .  \\
\end{aligned}
\end{equation}

To determine the transmit powers of each device, there are a variety of heuristics that could be employed. Two naive heuristics we consider are (i) allocating equal power to each transmitter and frequency band $P_{\text{max}} / 2T$ and (ii) randomly selecting half of the transmitters to communicate with power $P_{\text{max}} / T$ over each frequency band. 

Alternatively, we can use a learned policy to determine transmit powers, such as the parallel graph neural network, merged graph neural network, and multigraph neural network. As input for all learned policies, we use graph shift operators $\mathbf{B}^{(2.4)}$ and $\mathbf{B}^{(5)}$ and a signal $\mathbf{x}$ of all ones. Note that due to the fast fading channel coefficients, the underlying operators change with each timestep. All architectures contain two convolutional layers, each with a maximum diffusive order of 3, $G_\ell = 2$ filters, and a Sigmoid activation function (except for the last layer, which uses a ReLU). Pooling is not applied for any of the architectures.

To solve the constrained learning problem, we employ the primal-dual learning method detailed in \cite{9072356}. Primal updates are performed through the ADAM optimizer \cite{kingma2014adam}, and both primal and dual updates use a geometrically decaying step size. 

We consider a wireless system consisting of $40$ transmitters each assigned to one of $10$ receivers at random. Receivers are placed uniformly at random with positions $\mathbf{r}_i \in [-40,40]^2$ and associated transmitters placed randomly nearby at $\mathbf{t}_i \in [\mathbf{r}_i - 10, \mathbf{r}_i + 10]$. Unless otherwise specified, we select the additive white Gaussian noise $\omega^2 =1$ and $P_{\text{max}} = 100$mW.

For each training iteration, we consider a new random system configuration with 100 samples of fast fading channel coefficients. After 20,000 training iterations, we report the sum-rate on a validation set of 1,000 system configurations with 100 channel realizations each. 

In Fig. \ref{fig:powerResults}, we compare the performance of the heuristics and learned policies while varying the value of the power budget $P_{\text{max}}$ between 10mW, 50mW, and 100mW. In all three cases, the learned policies offer a considerable improvement over the heuristics, with the MGNN consistently yielding the highest sum-rate. The consideration of heterogeneous diffusions allows the MGNN to better allocate transmit power to maximize system throughput.

Fig. \ref{fig:noiseResults} showcases the advantage of the multigraph neural network in different SINR regimes. To vary the average SINR, we select the additive white Gaussian noise $\omega$ to range between $\{0.5,1,2\}\cdot 10^{-3}$mW, where a larger value of noise results in a lower average SINR. In all three regimes, the MGNN outperforms both of the graph neural networks. Though, as the level of noise in the environment increases, the discrepancies between the five architectures diminish.

\subsection{Hate Speech Spread on Twitter}

Hateful conduct on Twitter has been growing at an alarming rate \cite{siegel2021trumping,waseem2016hateful}. While many machine learning methods have been developed to classify whether an individual's tweet contains hateful rhetoric \cite{ayo2020machine,pariyani2021hate}, interventions applied against individual users are not particularly effective at containing hate speech. Rather, it has been found that hate speech often emerges from echo chambers \cite{masud2021hate}: communities of users whose hateful beliefs are amplified and broadcasted. Thus, to contain hate speech, we must identify the communities that enabled the hateful content to be spread, and apply strategic interventions at this level \cite{de2021multilingual, 10.1371/journal.pone.0265602, doi:10.1177/2056305120916850}.

This problem can be cast as a source localization task. Beginning with a tweet originating from one user, we observe the effect of the spreading process on other nodes at some time in the future. From this diffused signal, we seek to identify which community it originated from.


To construct our multigraph model, we use Twitter data from the 2015 Danish parliamentary election \cite{hanteer2019innovative}. In this election, there were 490 politicians running for parliament who were also active on Twitter. Each politician is associated with one of ten political parties, and one of the two main coalitions at the time\footnote{The red coalition contains the Alternativet, Enhedslisten, Radikale Venstre, Socialdemokratiet, and Socialistisk Folkeparti parties, while the blue coalition encompasses the Dansk Folkeparti, Det Konservative Folkeparti, KristenDemokraterne, Liberal Alliance, and Venstre parties.}: the red block and the blue block. We use these block and party affiliations as the communities for our task.

Interactions were collected for the 30 days leading up to the election. The interactions we consider are follows and retweets. For the follow graph with shift operator $\mathbf{S}_{\text{Fol}}$, we add a directed edge from $i$ to $j$ if user $j$ follows user $i$. On the retweet graph with shift operator $\mathbf{S}_{\text{Ret}}$, we include a directed edge from $i$ to $j$ if user $j$ retweets user $i$ over the period. 

To carry out the source localization, for each sample, we select a user $i \in \mathcal{V}$ uniformly at random. The node signal is the zero vector, except for the value of node $i$ which receives value one. We apply $K$ diffusions to the signal where $K$ a randomly drawn integer between 1 and 5, and each diffusion is randomly selected between $\mathbf{S}_{\text{Fol}}$ and $\mathbf{S}_{\text{Ret}}$, simulating the spread of the tweet to other users' feeds.

If the diffused signal does not reach at least half of the nodes, it is discarded. Each valid sample is associated with the block or party of the source node. We generate 20,000 samples and use a 80\%/20\% train/test split.

For the source localization task, we compare the performance of the parallel GNN, merged GNN, and multigraph neural network. All architectures employ two convolutional layers with $G_\ell = 32$ filters each, a ReLU nonlinearity function, and a linear readout layer mapping to the number of classes (2 for the block assignment task and 10 for the party assignment task). No pooling is applied for any of the architectures. We minimize the cross-entropy loss using an ADAM optimizer over 10 epochs, and evaluate using the classification accuracy. Results are reported as the average and standard deviation of the evaluation metric performance across 10 random splits of the dataset.

In the block source localization task, we seek to learn a mapping from the diffused signal to either the red block or the blue block, corresponding to the block affiliation of the source node. Results from this task are presented in Table \ref{table:blockResults}.


\begin{table}[h]
\centering
\begin{tabular}{lc}
\hline
Architecture        & Accuracy              \\ \hline
Parallel Graph Neural Network   & 91.2\% (±0.4)         \\
Merged Graph Neural Network   & 92.9\% (±0.6)         \\
\textbf{Multigraph Neural Network}   & \textbf{95.6\% }(±0.7) \\ \hline
\end{tabular}
\caption{Results of the Block Classification Task}
\label{table:blockResults}
\end{table}

All three learned architectures perform quite well on this task. However, the multigraph neural offers a significant boost in predictive power over both GNNs for this task. We hypothesize that the MGNN's inclusion of filters applied to heterogeneous diffusions allows the architecture to better model the complexities of information spread on a social network.

In the party source localization task, we instead use the diffused signal to learn to assign an affiliation of the source node to one of the ten parties. We present the performance of the three architectures in Table \ref{table:partyResults}.


\begin{table}[h]
\centering
\begin{tabular}{lc}
\hline
Architecture        & Accuracy              \\ \hline
Parallel Graph Neural Network     & 74.5\% (±1.8)   \\
Merged Graph Neural Network     & 77.5\% (±1.5)   \\
\textbf{Multigraph Neural Network}  & \textbf{86.6\%}  (±2.0)\\ \hline
\end{tabular}
\caption{Results of the Party Classification Task}
\label{table:partyResults}
\end{table}


On this more difficult task, similar results hold. The parallel and merged graph neural networks achieve modest classification accuracies of 74.5\% and 77.5\% respectively, while the multigraph neural network performs best at 86.6\%. We conclude that for this source localization task, the multigraph neural network offers a substantial performance improvement over the graph learning alternatives.



\section{Discussion and Conclusions}

We have introduced convolutional information processing on multigraphs, including the notions of signals, filtering, and frequency representation. This convolutional signal model was derived as a particular instantiation of a generic ASP model to perform scalable learning on multigraphs. Building on this, we proposed convolutional multigraph neural networks (MGNNs), enabling the processing of information supported on multiple graphs through learned filters.

We found that in some scenarios multigraph filters offer a richer representation than traditional graph filters. Additionally, we showed that the richer representations provided by multigraph filters are susceptible to a high computational cost that can be alleviated through operator pruning.

Compared to graph neural networks, which have been applied to the multiple graph setting through processing each graph in parallel, multigraph neural networks (MGNNs) better capture the complex dynamics that occur between and within each of the graphs. We provided numerical evidence of this claim, as the introduced architecture significantly outperforms graph neural networks on an optimal wireless resource allocation and a hate speech localization task.

One important property of the multigraph neural network is that its convolutions are parameterized by filters whose properties can be characterized by their spectral responses -- matrix polynomials in low dimensions. This makes multigraph filters transferable among multigraphs with equivalent spectral representations, i.e. the same number of frequencies and the same dimensions on the subspaces of frequencies. This raises the question if this attribute could lead to a form of transferability for MGNNs analogous to that developed for GNNs in~\cite{ruiz2020graphon}.

Under the introduced framework, operators are pruned through exploiting commutativity without usage of any training samples; instead, one could identify the most critical operators through conducting a sensitivity analysis using a small, held out set of samples.  In a similar vein, our selection procedure samples nodes through the use of heuristics, but could instead use coarsening or spectral-proxy methods designed for multigraphs. Improvements on these procedures remain interesting open areas for future research.

Additionally, given the representation of multigraph filters in~\eqref{eq_H_decomp_GM}, it is possible to see that their stability bounds can be studied in terms of $\bbH_i (\bbS_i)$ (where information is processed on individual graphs) and $\bbH_{[1:m]}(\bbS_1, \ldots, \bbS_m)$  (where information is processed mixing information between edges). This opens up an interesting research question about the role and impact of perturbations in $\bbH_i (\bbS_i)$ and $\bbH_{[1:m]}(\bbS_1, \ldots, \bbS_m)$. It will be important to better understand the trade-off between stability and selectivity when different restrictions -- different Lipschitz constants -- are imposed on each component in~\eqref{eq_H_decomp_GM}. From~\cite{algnn_nc_j}, we know that stability to deformations is guaranteed as long as the filters are Lipschitz and integral Lipschitz; however, it is not clear how to optimally reduce the values of those stability constants while preserving selectivity, particularly when a perturbation affects $\bbH_{[1:m]}(\bbS_1, \ldots, \bbS_m)$ more significantly than $\bbH_i (\bbS_i)$.


\bibliography{./bibliography}

\begin{thebibliography}{10}

\bibitem{sporns2022graph}
O.~Sporns.
\newblock Graph theory methods: applications in brain networks.
\newblock {\em Dialogues in clinical neuroscience}, 2022.

\bibitem{dijkstra2019networks}
H.A. Dijkstra, E.~Hern{\'a}ndez-Garc{\'\i}a, C.~Masoller, and M.~Barreiro.
\newblock {\em Networks in climate}.
\newblock Cambridge University Press, 2019.

\bibitem{bandyopadhyay2010rewiring}
S.~Bandyopadhyay, M.~Mehta, D.~Kuo, M.~Sung, R.~Chuang, E.J. Jaehnig,
  B.~Bodenmiller, K.~Licon, W.~Copeland, M.~Shales, et~al.
\newblock Rewiring of genetic networks in response to {DNA} damage.
\newblock {\em Science}, 330(6009):1385--1389, 2010.

\bibitem{meng2016interplay}
L.~Meng, Y.~Hulovatyy, A.~Striegel, and T.~Milenkovi{\'c}.
\newblock On the interplay between individuals’ evolving interaction patterns
  and traits in dynamic multiplex social networks.
\newblock {\em IEEE Transactions on Network Science and Engineering},
  3(1):32--43, 2016.

\bibitem{9072356}
M.~Eisen and A.~Ribeiro.
\newblock Optimal wireless resource allocation with random edge graph neural
  networks.
\newblock {\em IEEE Transactions on Signal Processing}, 68:2977--2991, 2020.

\bibitem{PhysRevLett.110.028701}
S.~G\'omez, A.~D\'{\i}az-Guilera, J.~G\'omez-Garde\~nes, C.~J. P\'erez-Vicente,
  Y.~Moreno, and A.~Arenas.
\newblock Diffusion dynamics on multiplex networks.
\newblock {\em Phys. Rev. Lett.}, 110:028701, Jan 2013.

\bibitem{zhang2019graph}
S.~Zhang, H.~Tong, J.~Xu, and R.~Maciejewski.
\newblock Graph convolutional networks: a comprehensive review.
\newblock {\em Computational Social Networks}, 6(1):1--23, 2019.

\bibitem{wang2020abstract}
D.~Wang, M.~Jamnik, and P.~Lio.
\newblock Abstract diagrammatic reasoning with multiplex graph networks.
\newblock {\em arXiv preprint arXiv:2006.11197}, 2020.

\bibitem{10.1145/3511808.3557572}
A.~Behrouz and F.~Hashemi.
\newblock Cs-mlgcn: Multiplex graph convolutional networks for community search
  in multiplex networks.
\newblock In {\em Proceedings of the 31st ACM International Conference on
  Information and Knowledge Management}, CIKM22, page 3828–3832, New York,
  NY, USA, 2022. Association for Computing Machinery.

\bibitem{9564196}
F.~Gao, Z.~Wang, and Z.~Liu.
\newblock Parallel multi-graph convolution network for metro passenger volume
  prediction.
\newblock In {\em 2021 IEEE 8th International Conference on Data Science and
  Advanced Analytics (DSAA)}, pages 1--10, 2021.

\bibitem{ke2021joint}
J.~Ke, S.~Feng, Z.~Zhu, H.~Yang, and J.~Ye.
\newblock Joint predictions of multi-modal ride-hailing demands: A deep
  multi-task multi-graph learning-based approach.
\newblock {\em Transportation Research Part C: Emerging Technologies},
  127:103063, 2021.

\bibitem{wang2021forecasting}
Z.~Wang, T.~Xia, R.~Jiang, X.~Liu, K.~Kim, Xuan Song, and R.~Shibasaki.
\newblock Forecasting ambulance demand with profiled human mobility via
  heterogeneous multi-graph neural networks.
\newblock In {\em 2021 IEEE 37th International Conference on Data Engineering
  (ICDE)}, pages 1751--1762. IEEE, 2021.

\bibitem{bertagnolli2021diffusion}
G.~Bertagnolli and M.~De~Domenico.
\newblock Diffusion geometry of multiplex and interdependent systems.
\newblock {\em Physical Review E}, 103(4):042301, 2021.

\bibitem{molina2015multiplex}
J.L. Molina-Borboa, S.~Mart{\'\i}nez-Jaramillo, F.~L{\'o}pez-Gallo, and
  M.~van~der Leij.
\newblock A multiplex network analysis of the mexican banking system: link
  persistence, overlap and waiting times.
\newblock {\em Journal of Network Theory in Finance}, 1(1):99--138, 2015.

\bibitem{lim2019discordant}
S.~Lim, F.~Radicchi, M.P. van~den Heuvel, and O.~Sporns.
\newblock Discordant attributes of structural and functional brain connectivity
  in a two-layer multiplex network.
\newblock {\em Scientific Reports}, 9(1):1--13, 2019.

\bibitem{huang2019principles}
Y.~Huang, H.~Dai, and R.~Ke.
\newblock Principles of effective and robust innate immune response to viral
  infections: a multiplex network analysis.
\newblock {\em Frontiers in immunology}, 10:1736, 2019.

\bibitem{zhang2018scalable}
H.~Zhang, L.~Qiu, L.~Yi, and Y.~Song.
\newblock Scalable multiplex network embedding.
\newblock In {\em IJCAI}, volume~18, pages 3082--3088, 2018.

\bibitem{qu2017attention}
M.~Qu, J.~Tang, J.~Shang, X.~Ren, M.~Zhang, and J.~Han.
\newblock An attention-based collaboration framework for multi-view network
  representation learning.
\newblock In {\em Proceedings of the 2017 ACM on Conference on Information and
  Knowledge Management}, pages 1767--1776, 2017.

\bibitem{fu2020magnn}
X.~Fu, J.~Zhang, Z.~Meng, and I.~King.
\newblock Magnn: Metapath aggregated graph neural network for heterogeneous
  graph embedding.
\newblock In {\em Proceedings of The Web Conference 2020}, pages 2331--2341,
  2020.

\bibitem{cen2019representation}
Y.~Cen, X.~Zou, J.~Zhang, H.~Yang, J.~Zhou, and J.~Tang.
\newblock Representation learning for attributed multiplex heterogeneous
  network.
\newblock In {\em Proceedings of the 25th ACM SIGKDD International Conference
  on Knowledge Discovery \& Data Mining}, pages 1358--1368, 2019.

\bibitem{yu2022multiplex}
P.~Yu, C.~Fu, Y.~Yu, C.~Huang, Z.~Zhao, and J.~Dong.
\newblock Multiplex heterogeneous graph convolutional network.
\newblock In {\em Proceedings of the 28th ACM SIGKDD Conference on Knowledge
  Discovery and Data Mining}, pages 2377--2387, 2022.

\bibitem{10.1145/3292500.3330961}
C.~Zhang, D.~Song, C.~Huang, A.~Swami, and N.V. Chawla.
\newblock Heterogeneous graph neural network.
\newblock In {\em Proceedings of the 25th ACM SIGKDD International Conference
  on Knowledge Discovery and Data Mining}, KDD '19, page 793–803, New York,
  NY, USA, 2019. Association for Computing Machinery.

\bibitem{aleta2017multilayer}
A.~Aleta, S.~Meloni, and Y.~Moreno.
\newblock A multilayer perspective for the analysis of urban transportation
  systems.
\newblock {\em Scientific reports}, 7(1):1--9, 2017.

\bibitem{natera2020extracting}
L.~Natera, F.~Battiston, G.~I{\~n}iguez, and M.~Szell.
\newblock Extracting the multimodal fingerprint of urban transportation
  networks.
\newblock {\em arXiv preprint arXiv:2006.03435}, 2020.

\bibitem{C2}
A.~Parada-Mayorga and A.~Ribeiro.
\newblock Algebraic neural networks: Stability to deformations.
\newblock {\em IEEE Transactions on Signal Processing}, 69:3351--3366, 2021.

\bibitem{algSP0}
M.~Püschel and J.~M.~F. Moura.
\newblock Algebraic signal processing theory: An overview, 2006.

\bibitem{algSP1}
M.~{Puschel} and J.~M.~F. {Moura}.
\newblock Algebraic signal processing theory: Foundation and 1-d time.
\newblock {\em IEEE Transactions on Signal Processing}, 56(8):3572--3585, Aug
  2008.

\bibitem{algnn_nc_j}
A.~Parada-Mayorga, L.~Butler, and A.~Ribeiro.
\newblock Convolutional filtering and neural networks with non commutative
  algebras.
\newblock {\em CoRR}, abs/2108.09923, 2021.

\bibitem{jointblock1}
K.~Abed-Meraim and A.~Belouchrani.
\newblock Algorithms for joint block diagonalization.
\newblock In {\em 2004 12th European Signal Processing Conference}, pages
  209--212, 2004.

\bibitem{jointblock2}
I.~Bischer, C.~Döring, and A.~Trautner.
\newblock Simultaneous block diagonalization of matrices of finite order.
\newblock {\em Journal of Physics A: Mathematical and Theoretical},
  54(8):085203, feb 2021.

\bibitem{ortega_gsp}
A.~{Ortega}, P.~{Frossard}, J.~{Kovačević}, J.~M.~F. {Moura}, and
  P.~{Vandergheynst}.
\newblock Graph signal processing: Overview, challenges, and applications.
\newblock {\em Proceedings of the IEEE}, 106(5):808--828, May 2018.

\bibitem{algSP2}
M.~{Puschel} and J.~M.~F. {Moura}.
\newblock Algebraic signal processing theory: 1-d space.
\newblock {\em IEEE Transactions on Signal Processing}, 56(8):3586--3599, Aug
  2008.

\bibitem{algSP3}
J.~{Kovacevic} and M.~{Puschel}.
\newblock Algebraic signal processing theory: Sampling for infinite and finite
  1-d space.
\newblock {\em IEEE Transactions on Signal Processing}, 58(1):242--257, Jan
  2010.

\bibitem{repthybigbook}
M.~Lorenz.
\newblock {\em A Tour of Representation Theory}.
\newblock Graduate studies in mathematics. American Mathematical Society, 2018.

\bibitem{repthysmbook}
P.I. Etingof, O.~Golberg, S.~Hensel, T.~Liu, A.~Schwendner, D.~Vaintrob, and
  E.~Yudovina.
\newblock {\em Introduction to Representation Theory}.
\newblock Student mathematical library. American Mathematical Society, 2011.

\bibitem{9206091}
F.~Gama, J.~Bruna, and A.~Ribeiro.
\newblock Stability properties of graph neural networks.
\newblock {\em IEEE Transactions on Signal Processing}, 68:5680--5695, 2020.

\bibitem{defferrard2016convolutional}
M.~Defferrard, X.~Bresson, and P.~Vandergheynst.
\newblock Convolutional neural networks on graphs with fast localized spectral
  filtering.
\newblock {\em Advances in neural information processing systems}, 29, 2016.

\bibitem{gama2018convolutional}
F.~Gama, A.G. Marques, G.~Leus, and A.~Ribeiro.
\newblock Convolutional neural network architectures for signals supported on
  graphs.
\newblock {\em IEEE Transactions on Signal Processing}, 67(4):1034--1049, 2018.

\bibitem{9244194}
D.L. Lau, G.R. Arce, A.~Parada-Mayorga, D.~Dapena, and K.~Pena-Pena.
\newblock Blue-noise sampling of graph and multigraph signals: Dithering on
  non-euclidean domains.
\newblock {\em IEEE Signal Processing Magazine}, 37(6):31--42, 2020.

\bibitem{chen2015discrete}
S.~Chen, R.~Varma, A.~Sandryhaila, and J.~Kova{\v{c}}evi{\'c}.
\newblock Discrete signal processing on graphs: Sampling theory.
\newblock {\em IEEE transactions on signal processing}, 63(24):6510--6523,
  2015.

\bibitem{anis2016efficient}
A.~Anis, A.~Gadde, and A.~Ortega.
\newblock Efficient sampling set selection for bandlimited graph signals using
  graph spectral proxies.
\newblock {\em IEEE Transactions on Signal Processing}, 64(14):3775--3789,
  2016.

\bibitem{tsitsvero2016signals}
M.~Tsitsvero, S.~Barbarossa, and P.~Di~Lorenzo.
\newblock Signals on graphs: Uncertainty principle and sampling.
\newblock {\em IEEE Transactions on Signal Processing}, 64(18):4845--4860,
  2016.

\bibitem{puy2018random}
G.~Puy, N.~Tremblay, R.~Gribonval, and P.~Vandergheynst.
\newblock Random sampling of bandlimited graph signals.
\newblock {\em Appl. Comput. Harmonic Anal}, 44(2):446--475, 2018.

\bibitem{varma2015spectrum}
R.~Varma, S.~Chen, and J.~Kova{\v{c}}evi{\'c}.
\newblock Spectrum-blind signal recovery on graphs.
\newblock In {\em 2015 IEEE 6th International Workshop on Computational
  Advances in Multi-Sensor Adaptive Processing (CAMSAP)}, pages 81--84. IEEE,
  2015.

\bibitem{https://doi.org/10.48550/arxiv.2006.13318}
C.~Cai and Y.~Wang.
\newblock A note on over-smoothing for graph neural networks, 2020.

\bibitem{liu2012achieving}
L.~Liu, R.~Zhang, and K.C. Chua.
\newblock Achieving global optimality for weighted sum-rate maximization in the
  k-user gaussian interference channel with multiple antennas.
\newblock {\em IEEE Transactions on Wireless Communications}, 11(5):1933--1945,
  2012.

\bibitem{8821330}
Y.~Li, S.~Xia, M.~Zheng, B.~Cao, and Q.~Liu.
\newblock Lyapunov optimization-based trade-off policy for mobile cloud
  offloading in heterogeneous wireless networks.
\newblock {\em IEEE Transactions on Cloud Computing}, 10(1):491--505, 2022.

\bibitem{4275017}
M.~Chiang, C.W. Tan, D.P. Palomar, D.~O'neill, and D.~Julian.
\newblock Power control by geometric programming.
\newblock {\em IEEE Transactions on Wireless Communications}, 6(7):2640--2651,
  2007.

\bibitem{chen2011round}
C.S. Chen, K.W. Shum, and C.W. Sung.
\newblock Round-robin power control for the weighted sum rate maximisation of
  wireless networks over multiple interfering links.
\newblock {\em European Transactions on Telecommunications}, 22(8):458--470,
  2011.

\bibitem{shi2011iteratively}
Q.~Shi, M.~Razaviyayn, Z.Q. Luo, and C.~He.
\newblock An iteratively weighted mmse approach to distributed sum-utility
  maximization for a mimo interfering broadcast channel.
\newblock {\em IEEE Transactions on Signal Processing}, 59(9):4331--4340, 2011.

\bibitem{liang2019towards}
F.~Liang, C.~Shen, W.~Yu, and F.~Wu.
\newblock Towards optimal power control via ensembling deep neural networks.
\newblock {\em IEEE Transactions on Communications}, 68(3):1760--1776, 2019.

\bibitem{lima2020resource}
V.~Lima, M.~Eisen, K.~Gatsis, and A.~Ribeiro.
\newblock Resource allocation in large-scale wireless control systems with
  graph neural networks.
\newblock {\em IFAC-PapersOnLine}, 53(2):2634--2641, 2020.

\bibitem{goldsmith2005wireless}
A.~Goldsmith.
\newblock {\em Wireless communications}.
\newblock Cambridge university press, 2005.

\bibitem{kingma2014adam}
D.P. Kingma and J.~Ba.
\newblock Adam: A method for stochastic optimization.
\newblock {\em arXiv preprint arXiv:1412.6980}, 2014.

\bibitem{siegel2021trumping}
A.A. Siegel, E.~Nikitin, P.~Barber{\'a}, J.~Sterling, B.~Pullen, R.~Bonneau,
  J.~Nagler, J.A. Tucker, et~al.
\newblock Trumping hate on {T}witter? {O}nline hate speech in the 2016 us
  election campaign and its aftermath.
\newblock {\em Quarterly Journal of Political Science}, 16(1):71--104, 2021.

\bibitem{waseem2016hateful}
Z.~Waseem and D.~Hovy.
\newblock Hateful symbols or hateful people? {P}redictive features for hate
  speech detection on {T}witter.
\newblock In {\em Proceedings of the NAACL student research workshop}, pages
  88--93, 2016.

\bibitem{ayo2020machine}
F.E. Ayo, O.~Folorunso, F.T. Ibharalu, and I.A. Osinuga.
\newblock Machine learning techniques for hate speech classification of
  {T}witter data: State-of-the-art, future challenges and research directions.
\newblock {\em Computer Science Review}, 38:100311, 2020.

\bibitem{pariyani2021hate}
B.~Pariyani, K.~Shah, M.~Shah, T.~Vyas, and S.~Degadwala.
\newblock Hate speech detection in {T}witter using natural language processing.
\newblock In {\em 2021 Third International Conference on Intelligent
  Communication Technologies and Virtual Mobile Networks (ICICV)}, pages
  1146--1152. IEEE, 2021.

\bibitem{masud2021hate}
S.~Masud, S.~Dutta, S.~Makkar, C.~Jain, V.~Goyal, A.~Das, and T.~Chakraborty.
\newblock Hate is the new infodemic: A topic-aware modeling of hate speech
  diffusion on {T}witter.
\newblock In {\em 2021 IEEE 37th International Conference on Data Engineering
  (ICDE)}, pages 504--515. IEEE, 2021.

\bibitem{de2021multilingual}
G.L. De~La Pe{\~n}a~Sarrac{\'e}n.
\newblock Multilingual and multimodal hate speech analysis in {T}witter.
\newblock In {\em Proceedings of the 14th ACM International Conference on Web
  Search and Data Mining}, pages 1109--1110, 2021.

\bibitem{10.1371/journal.pone.0265602}
B.~Evkoski, A.~Pelicon, I.~Mozetič, N.~Ljubešić, and P.~Kralj~Novak.
\newblock Retweet communities reveal the main sources of hate speech.
\newblock {\em PLOS ONE}, 17(3):1--22, 03 2022.

\bibitem{doi:10.1177/2056305120916850}
S.~Ozalp, M.L. Williams, P.~Burnap, H.~Liu, and M.~Mostafa.
\newblock Antisemitism on {T}witter: Collective efficacy and the role of
  community organisations in challenging online hate speech.
\newblock {\em Social Media + Society}, 6(2):2056305120916850, 2020.

\bibitem{hanteer2019innovative}
O.~Hanteer and L.~Rossi.
\newblock An innovative way to model {T}witter topic-driven interactions using
  multiplex networks.
\newblock {\em Frontiers in Big Data}, page~9, 2019.

\bibitem{ruiz2020graphon}
L.~Ruiz, L.F.O. Chamon, and A.~Ribeiro.
\newblock Graphon signal processing, 2020.

\end{thebibliography}
\bibliographystyle{unsrt}



\appendices
\section{Pruning Procedure of the Diffusion Tree}
\label{app:pruning}

In Algorithm~\ref{alg:Pruning}, we show the details of the pruning method applied during the generation of a depth-limited operator tree. The first step of the algorithm computes the commutator of all pairs of operators; if the commutator is less than a given cutoff parameter $\epsilon$, either $\mathbf{S}_i \mathbf{S}_j$ or $\mathbf{S}_j \mathbf{S}_i$ are arbitrarily chosen to be pruned. The second step is to recursively generate all combinations of operators up to a depth $k$, removing elements (and their corresponding branches) with indices contained in $\texttt{pruned}$. Lastly, the valid operator indices $\texttt{valInd}$ are used to construct the set of multigraph operators $\mathcal{Q}$.

\begin{figure*}[t]
\centering
\includegraphics[width=\textwidth]{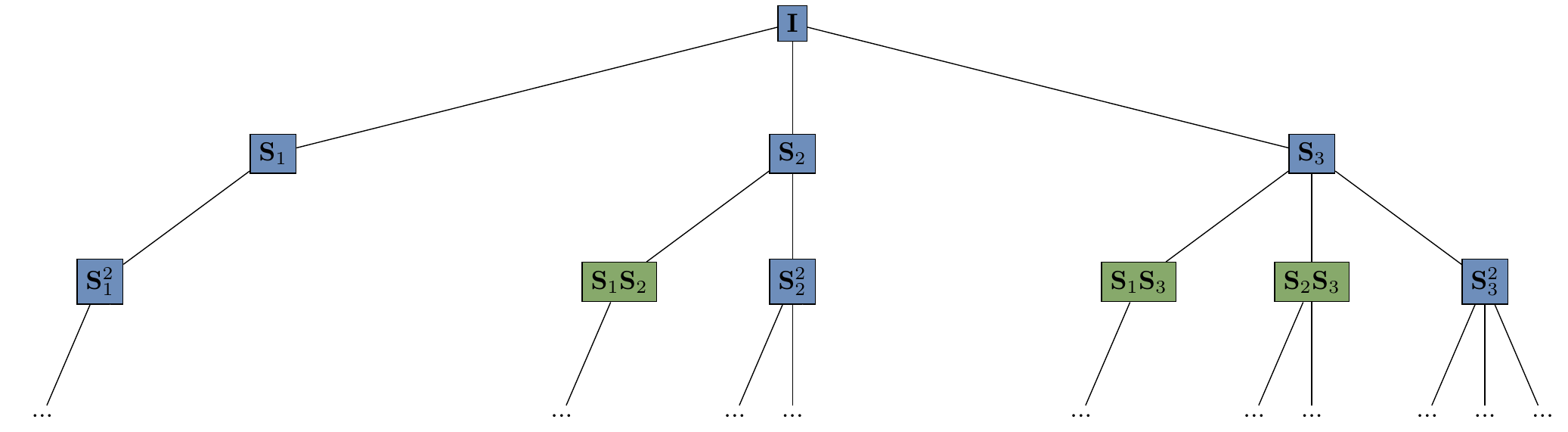}
\caption{Diffusion tree for operators $\mathbf{S}_1$, $\mathbf{S}_2$, and $\mathbf{S}_3$ where any term containing $\mathbf{S}_2 \mathbf{S}_1$, $\mathbf{S}_3 \mathbf{S}_1$, or $\mathbf{S}_3 \mathbf{S}_2$ is pruned.}
\label{fig:prunedDiffTree}
\end{figure*}

Figure~\ref{fig:prunedDiffTree} illustrates an example of the tree structure obtained as a result of Algorithm~\ref{alg:Pruning} in a multigraph diffusion tree with three shift operators where terms containing $\mathbf{S}_2 \mathbf{S}_1$, $\mathbf{S}_3 \mathbf{S}_1$, or $\mathbf{S}_3 \mathbf{S}_2$ are pruned.


\begin{algorithm}
    \caption{Generation of a Pruned, Depth-Limited Diffusion Tree}
    \label{alg:Pruning}
\begin{algorithmic}
\Require  $(\mathbf{S}_1, \dots, \mathbf{S}_m)$: shift operators, $\epsilon$: cutoff parameter, \\ $K$: tree depth
\Ensure $\mathcal{Q}$: set of multigraph operators
\LineComment{\emph{Determine terms to prune:}}
\State $\texttt{pruned}, \texttt{frontier}, \texttt{valInd} \gets \{\}, \{\}, \{\}$
\For{$i, j \gets 1$ to $m$ for $i < j$}
\State \texttt{frontier} $\gets$ \texttt{frontier}  $ \cup \hspace{0.5em} \{(i)\}$
\If{$|| \mathbf{S}_i\mathbf{S}_j - \mathbf{S}_j\mathbf{S}_i ||_2 \leq \epsilon$}
\State \texttt{pruned} $\gets$ \texttt{pruned}  $ \cup \hspace{0.5em} \{(j,i)\}$
\EndIf
\EndFor
\State 
\LineComment{\emph{Generate indices of valid terms:}}
\While{$|\texttt{frontier}| \hspace{0.25em}> 0$}
\State $tup \gets \texttt{frontier}_0$ \Comment{Extract any element}
\For{$k \gets 1$ to $m$}
\If{$(k,tup[0]) \notin$ \texttt{pruned} and $|tup| < K$}
\State \texttt{frontier} $\gets$ \texttt{frontier}  $ \cup \hspace{0.5em} \{(k) + tup\}$
\State \texttt{valInd} $\gets$ \texttt{valInd}  $ \cup \hspace{0.5em} \{(k) + tup\}$
\EndIf
\EndFor
\EndWhile

\State
\State Use $(\mathbf{S}_1, \dots, \mathbf{S}_m)$ to construct $\mathcal{Q}$ from \texttt{valInd}
\end{algorithmic}
\end{algorithm}


\section{Label Permutation Equivariance of Multigraph Neural Networks}
\label{app:permEqui}

\begin{proof} Exploiting the orthogonality of $\mathbf{P}$, it follows that all compositions of permuted shift operators are permutations of the same composition of shift operators

\begin{equation}
    \hat{\mathbf{S}}_i\dots \hat{\mathbf{S}}_j = \mathbf{P}^\intercal \mathbf{S}_i\mathbf{P}\dots \mathbf{P}^\intercal \mathbf{S}_j\mathbf{P}=    \mathbf{P}^\intercal\mathbf{S}_i\dots \mathbf{S}_j\mathbf{P}.
\end{equation}

We can substitute this fact into the definition of the permuted multigraph filter

\begin{equation}
\begin{split}
\mathbf{H}_\ell(\hat{\mathbf{S}}_1, \dots, \hat{\mathbf{S}}_m) \\
&\hspace{-1.8cm}= f_{\hat{\mathbf{I}}}\hat{\mathbf{I}} + f_{\hat{\mathbf{S}}_1}\hat{\mathbf{S}}_1 + \dots + f_{\hat{\mathbf{S}}_i\hat{\mathbf{S}}_j}\hat{\mathbf{S}}_i\hat{\mathbf{S}}_j + \dots \\ 
&\hspace{-1.8cm}= f_{\hat{\mathbf{I}}}\mathbf{P}^\intercal\mathbf{I}\mathbf{P} + f_{\hat{\mathbf{S}}_1}\mathbf{P}^\intercal\mathbf{S}_1\mathbf{P} + \dots + f_{\hat{\mathbf{S}}_i\hat{\mathbf{S}}_j}\mathbf{P}^\intercal\mathbf{S}_i\mathbf{S}_j\mathbf{P} + \dots \\
&\hspace{-1.8cm}= \mathbf{P}^\intercal\left(f_{\hat{\mathbf{I}}}\mathbf{I} + f_{\hat{\mathbf{S}}_1}\mathbf{S}_1 + \dots + f_{\hat{\mathbf{S}}_i\hat{\mathbf{S}}_j}\mathbf{S}_i\mathbf{S}_j + \dots\right)\mathbf{P} \\
&\hspace{-1.8cm}= \mathbf{P}^\intercal\mathbf{H}_\ell(\mathbf{S}_1, \dots, \mathbf{S}_m)\mathbf{P}
\end{split}
\end{equation}

Given that the nonlinearity $\sigma(\cdot)$ employed is pointwise, when applying the permuted filter to the permuted signal $\hat{\mathbf{x}}_{\ell - 1} = \mathbf{P}^\intercal\mathbf{x}_{\ell - 1}$, it follows
\begin{equation}
    \begin{split}
        \hat{\mathbf{x}}_\ell &= \sigma(\mathbf{H}_\ell(\hat{\mathbf{S}}_1, \dots, \hat{\mathbf{S}}_m)\hat{\mathbf{x}}_{\ell - 1}) \\
&= \sigma(\mathbf{P}^\intercal\mathbf{H}_\ell(\mathbf{S}_1, \dots, \mathbf{S}_m)\mathbf{P}\mathbf{P}^\intercal\mathbf{x}_{\ell - 1}) \\
&= \mathbf{P}^\intercal\sigma(\mathbf{H}_\ell(\mathbf{S}_1, \dots, \mathbf{S}_m)\mathbf{x}_{\ell - 1}) \\
&= \mathbf{P}^\intercal\mathbf{x}_\ell,
    \end{split}
\end{equation}
which holds for any $\ell = 1, \dots, L$, completing the proof. \end{proof}

\end{document}